\documentclass[10pt,journal,compsoc]{IEEEtran}
\usepackage{cite}
\usepackage[pdftex]{graphicx}
\usepackage{booktabs}
\usepackage{amsmath}
\usepackage{algorithmic}
\usepackage{caption}
\usepackage{url}
\usepackage{amsmath,amssymb,amsfonts}
\usepackage{amsthm}
\usepackage{algorithmic}
\usepackage{graphicx}
\usepackage{diagbox}
\usepackage{textcomp}
\usepackage{xcolor}
\usepackage{wrapfig}
\def\BibTeX{{\rm B\kern-.05em{\sc i\kern-.025em b}\kern-.08em
		T\kern-.1667em\lower.7ex\hbox{E}\kern-.125emX}}
\usepackage{subcaption}
\usepackage[ruled,vlined]{algorithm2e}
\usepackage{multirow}

\usepackage{soul}

\newtheorem{df}{Definition}

\DeclareMathOperator*{\argmax}{arg\,max}
\DeclareMathOperator*{\argmin}{arg\,min}

\begin{document}
	\title{FocusedCleaner: Sanitizing Poisoned Graphs \\for Robust GNN-based Node Classification}

	\author{Yulin~Zhu, %~\IEEEmembership{Member,~IEEE,}
		Liang~Tong, %~\IEEEmembership{Fellow,~OSA,}
		Gaolei~Li, %~\IEEEmembership{Life~Fellow,~IEEE}% <-this % stops a space
		Xiapu Luo, Kai Zhou
		}
	
\markboth{Journal of \LaTeX\ Class Files,~Vol.~14, No.~8, August~2015}%
{Shell \MakeLowercase{\textit{et al.}}: Bare Demo of IEEEtran.cls for Computer Society Journals}

\IEEEtitleabstractindextext{%
	\begin{abstract}
		Graph Neural Networks (GNNs) are vulnerable to data poisoning attacks, which will generate a poisoned graph as the input to the GNN models. We present FocusedCleaner as a poisoned graph sanitizer to effectively identify the poison injected by attackers. Specifically, FocusedCleaner provides a sanitation framework consisting of two modules: bi-level structural learning and victim node detection. In particular, the structural learning module will reverse the attack process to steadily sanitize the graph while the detection module provides the ``focus” -- a narrowed and more accurate search region -- to structural learning. These two modules will operate in iterations and reinforce each other to sanitize a poisoned graph step by step. As an important application, we show that the adversarial robustness of GNNs trained over the sanitized graph for the node classification task is significantly improved. Extensive experiments demonstrate that FocusedCleaner outperforms the state-of-the-art baselines both on poisoned graph sanitation and improving robustness.
	\end{abstract}
	
	}

\maketitle
\IEEEdisplaynontitleabstractindextext
\IEEEpeerreviewmaketitle

%%%%%%%%%%%%%%%%%%%%%%%%%%%%%%%%%%%%%%%%%%%
% Add sections here
%%%%%%%%%%%%%%%%%%%%%%%%%%%%%%%%%%%%%%%%%%

\IEEEraisesectionheading{\section{Introduction}\label{sec:introduction}}
In the past few years, Graph Neural Networks (GNNs) are potent deep learning models to capture semantic and structural information of the relational data for widely-adopted downstream tasks such as node classification~\cite{gcn, graphsage}, link prediction~\cite{LinkPred}, community detection~\cite{community}, graph classification~\cite{GIN} and graph anomaly detection~\cite{GADsurvey} due to their specially-designed aggregation mechanism. The aggregation mechanism can propagate the node's features along the topology on the non-Euclidean space and obtain the node representations with good quality. However, extensive research efforts have been devoted to studying the robustness of GNNs under attack and verifying that GNNs are prone to be attacked by imperceptible structural attacks~\cite{mettack, topologyattack}. In particular, a prevalent type of attack, termed \textit{data poisoning attack}~\cite{poison}, studies how to manipulate the input graph data (especially the graph structure) to mislead GNNs' predictions. Basically, an attacker can tamper with the data collection process to create a \textit{poisoned graph}, over which an analyst runs GNN-based prediction algorithms that would make 
possibly wrong predictions. Such poisoning attacks are shown to be very effective in various graph analytic tasks~\cite{mettack,topologyattack,binarizedattack}.

% problem, importance
Against this background, we are thus motivated to investigate this important problem of \textbf{poisoned graph sanitation}: given a poisoned graph, how can the analyst (defender) identify the \textit{poison} injected by the attacker? This problem is important for a few reasons. First, unlike attacks over images where the poison is meaningless noise, poison over graphs (e.g., edges and nodes) often possess some physical meanings. Thus, identifying the poison itself could be useful. For example, consider a poisoning attack against a recommendation system~\cite{RS} where the attacker modifies the \textit{user-item} bipartite graph. The poison injected by the attacker corresponds to the actual reviews, the successful identification of which could help to locate the fake/compromised user accounts and further trace the attacker. Second, data sanitation can act as a data preprocessing step, which can serve as an important defense mechanism -- indeed, we demonstrate that our sanitation techniques can greatly enhance the robustness of GNN-based node classification. In this paper, we focus on sanitizing the structural noise (i.e., edges) introduced by attackers.

Graph sanitation problem resembles a well-investigated topic termed \textit{graph structural learning}. Specifically, given a graph $\mathcal{G}$ contaminated by noise (could be random or adversarial), graph structural learning aims to introduce some \textit{slight refinement} to the graph structure (thus learning a new graph $\mathcal{G}'$) in the hope that the refinement can ``neutralize” the  effects of the noise so as to improve model performance.  
For instance, a representative line of research applies data augmentation techniques \cite{aug,localaug,GASOLINE} to boost the node classification accuracy of GNNs. In particular, the work closely related to ours is \textit{GASOLINE} \cite{GASOLINE}, which utilizes a bi-level structural learning framework to search for a ``better” topology to increase node classification accuracy.
However, we emphasize that structural learning and data sanitation have very distinct goals. The formal aims to improve model performance by learning a new graph $\mathcal{G}'$ with no intention to recover the original clean graph $\mathcal{G}_0$.
%does not intend to distinguish genuine data from poison. In fact, the newly learned graph is not necessarily similar to the original clean graph -- it is just that the learned graph can result in better model performance. 
In comparison, sanitation explicitly aims to \textbf{identify and remove the poison} to recover $\mathcal{G}_0$, and importantly a natural result of sanitation is improved model performance. We elaborate on this key difference with supporting experiment results in Section~\ref{sec-rethinking}.

 %Thus, it prompts us to \textbf{rethink the graph sanitation problem}: Identifying and sanitizing the adversarial noises is more powerful than ``neutralizing” the malicious effect caused by them (to be detailed later).

%The preprocessing based defense method belongs to a broader topic: graph structural learning. For the vast majority of existing works, they essentially aim at mining and updating the topology information of the social networks to enhance the performance of specific tasks or to recover the intrinsic properties of the graph, like homophily \cite{homophily}. A common knowledge of this field is the initially given graph has already been polluted by noisy links or nodes, or the missing links or nodes due to the perfunctory operation during the data collection phase. The mining model endeavors to mitigate or ``neutralize” the deficiency of the graph by inserting or deleting the links, nodes or manipulating the node attributes. For example, flourish literatures have worked on implementing the data augmentation \cite{aug, localaug} to the relational data to boost the node classification performance on GNN. It is worth noting that those data augmentation models utilizes the prior knowledge that by removing the inter-class links and adding the intra-class links can boost the node classification performance. In fact, augmenting the prior knowledge of a specific scenario will play a key point in this task, thus approximate to a ``better” graph. 

% introduce our work
To tackle the problem of poisoned graph sanitation, we present \textsf{FocusedCleaner}, a framework composed of two modules: a bi-level structural learning module and a victim node detection module. Importantly, these two modules will operate collaboratively to enhance each other. On the one hand, the detection module acts as a supervisor and provides the ``focus” to narrow down the search region for the structural learning module, which can more accurately pick out the adversarial links. On the other hand, 
%the input graph with its GNN features obtained from the inner training of the bi-level structural learning module will provide affluent information for victim node detection. 
the inner training component of the structural learning module can learn node features with higher quality, which boosts the performance of the detection module.
Overall, these two modules will reinforce each other and provide better sanitation results.

The application of \textsf{FocusedCleaner} to enhance the adversarial robustness of GNN-based node classification is immediate: GNN models trained over the sanitized graph can achieve much better performance. That is, \textsf{FocusedCleaner} can effectively serve as a preprocessing-based defense approach against data poisoning attacks. 

In summary, we propose a poisoned graph sanitizer \textsf{FocusedCleaner} that possesses the following nice features:
\begin{itemize}
    \item \textsf{FocusedCleaner} can achieve better sanitation performance than state-of-the-art baseline approaches in terms of effectively identifying the attacker-injected edges in a poisoned graph to recover the clean graph.
    \item To effectively validate the poisoned graph sanitation performance of the preprocessing-based defense methods, we introduce a rational metric--$\mathsf{ESR}$ based on the Jaccard index~\cite{jaccardidx}. Higher $\mathsf{ESR}$ will probably lead to better robustness for node classification. Sanitation evaluation based on a traditional metric like $F1$ score is also given and achieve the consistent performance with $\mathsf{ESR}$. 
    \item GNN models trained over graphs sanitized by \textsf{FocusedCleaner} have higher node classification accuracy compared to training over graphs sanitized by other graph structural learning methods. That is, \textsf{FocusedCleaner} can provide better defense performance.
    \item As a preprocessing-based defense approach, \textsf{FocusedCleaner} outperforms state-of-the-art robust GNN models. Moreover, \textsf{FocusedCleaner} can also be used in combination with robust GNN models, further boosting defense performance.
\end{itemize}
\begin{figure*}
	\centering
    \includegraphics[width=\textwidth,height=3.8cm]{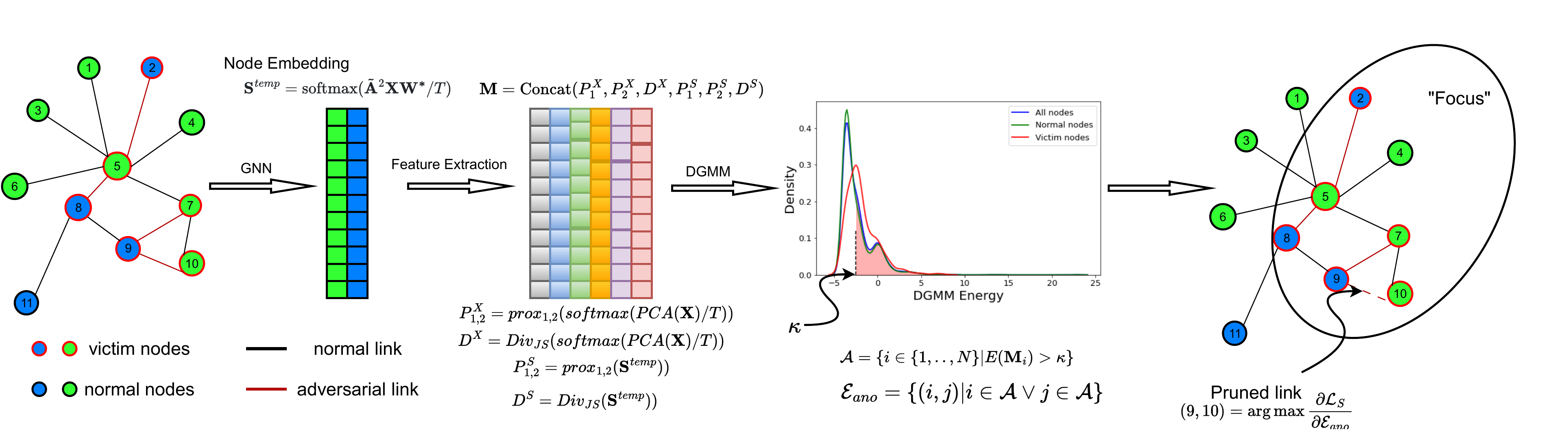}
	\caption{Overall framework of the \textsf{FocusedCleaner}.}
	\label{fig-focusedcleaner-framework}
\end{figure*}

\section{Related Works}
%\mynote{fix the reference style}
\subsection{Graph Structural Learning}
%\mynote{citation style}
Graph structural learning aims to learn a slightly different graph topology to boost the performance of a specific graph learning task, such as node classification~\cite{gcn, graphsage}, link prediction~\cite{LinkPred}, community detection~\cite{community}, etc. They assume the initially given graph has been contaminated by the noisy links or nodes, and aim at mining the topology of the graph for better downstream tasks' performances. For example, \cite{aug} utilizes the GVAE \cite{VGAE} to prune the inter-cluster links and insert the intra-cluster links and train the GNN with the refined graph. \cite{localaug} augment the graph locally by sampling neighbors' GNN features to prevent the degenerated performance of the aggregation mechanism when the number of neighbors is limited. \cite{MH} augments the graph's topology information by MH algorithm~\cite{MHalgorithm} which can well capture the distribution of the graph structure. \cite{NE} removes the weak links and enhances real connections to the biological networks by using a
doubly stochastic matrix operator and increasing the spectrum eigengap of the input graph. \cite{maskGVAE} iteratively trains the min-cut loss~\cite{mincut} and GVAE and removes the irrelevant edges. \cite{Enet} leverages the mutual influence of noisy links detection and missing links prediction to enhance each other. All the above methods endeavor to find a better graph structure to enhance the performance of the downstream tasks.

\subsection{Graph Defense}
Graph defense methods aim at defending against structural poisoning or evasion attacks on the graph data and try to preserve the node classification performance under varying attacking powers. It contains two types: preprocessing-based methods and robust models. For preprocessing-based methods, \cite{gcnjaccard} utilizes the Jaccard similarity to prune the potential adversarial links. \cite{GASOLINE} implements a bi-level optimization to enhance the robustness of GNN. For robust models, \cite{RGCN} learns a Gaussian distribution on node features and prunes the nodes with large variance. \cite{GNNGUARD} adopted the cosine similarity between two nodes' GNN features and obtain the link dropout probability during training. \cite{simpgcn} constructed a kNN graph based on node attributes and incorporate the kNN graph into GNN to ensure attribute consistency. \cite{mediangnn} dedicated to mitigating the influence of bad propagation by introducing robust statistics. \cite{elasticgnn} implemented the $L1$-norm to the graph signal estimator to enhance the robustness of the model. \cite{airgnn} introduced an adaptive residual mechanism to mitigate the abnormal node features' impact.
%\subsection{Problem Statements}
%In this section, we introduce the necessary preliminaries on GNN-based node classification and structural poisoning attacks, and then define the problem of graph data sanitization.
\section{Preliminaries}
\subsection{Notations}
In this section, we first introduce the frequently used notations in Tab.~\ref{tab-notations}. Next, we give a brief introduction to graph neural network and structural poisoning attack.

\begin{table}[h]
    \centering
    \caption{Frequently used notations.}
    \label{tab-notations}
    \begin{tabular}{c|c}
    \toprule[1.pt]
    Notations & Descriptions \\
    \hline
    $N$ & node number \\
    $E$ & link set of the graph \\
    $\mathcal{G}^{p}$ & Given poisoned graph \\
    $\mathcal{G}^{R}$ & A sanitized graph recovered from $\mathcal{G}^{p}$ \\
    $\mathbf{A}^{p}$ & Adjacency matrix of a poisoned graph \\
    $\mathbf{X}$ & Nodal attributes matrix of a graph \\
    $\mathbf{D}^{p}$ & Degree matrix of a poisoned graph \\
    $B$ & Budget of the attacker \\
    $\mathbf{W}$ & Learnable weight matrix of GNN \\
    $\mathbf{Y}$ & Label matrix of a graph \\
    $\mathcal{A}$ & Victim nodes set \\
    $\mathcal{N}$ & Normal nodes set \\
    $\mathbf{L}^{p}$ & Laplacian matrix of a poisoned graph \\
    $\lambda_{\mathcal{V}}$ & relative importance of the validation loss \\
    $\lambda_{\mathcal{T}^{\prime}}$ & relative importance of the testing loss \\
    $\mathbf{S}$ & output of a two-layered GNN \\
    \bottomrule[1.pt]
    \end{tabular}
\end{table}

%We frame the problem of graph data sanitization in the context of semi-supervised node classification since it is one of the most prevalent tasks for GNNs, and many data poisoning attacks also target node classification.
\subsection{GNN-Based Node Classification} 
%Recently, GNNs have become the \textit{de facto} choice for a wide range of analytic tasks over graph data. In this paper, we focus on the node classification task as a representative. We consider a typical transductive setting where, 
We frame the problem of graph data sanitization in the context of semi-supervised node classification. 
%Given a graph 
%with partially labeled nodes, the task of node classification is to predict the labels for the remaining nodes. 
Specifically, denote a graph as 
$\mathcal{G} = (\mathbf{V}_l, \mathbf{V}_u, \mathbf{X}, \mathbf{A}, \mathbf{Y})$, where $\mathbf{V}_l$/$\mathbf{V}_u$ are the sets of labeled/unlabeled nodes, $\mathbf{Y}$ denotes the set of available node labels, $\mathbf{X}$ is the attribute matrix and $\mathbf{A}$ is the adjacent matrix.  A GNN model~\cite{gcn} denoted as  $f_\theta(\mathcal{G}) \rightarrow \{y_i \in \mathcal{C}, \forall v_i \in \mathbf{V}_u\}$ then can be trained to predict the missing  node labels, where $\theta$ summarizes the trainable model parameters, $\mathcal{C}$ is a finite set of labels, $y_i$ is the predicted label for a node $v_i$ in $\mathbf{V}_u$. While there are various GNN models, a typical construction of the GNN layers \cite{gcn} is as follows:
\begin{equation}
\begin{split}
    &\mathbf{H}^{(l+1)}=\sigma(\mathbf{\tilde{D}}^{-\frac{1}{2}}\tilde{\mathbf{A}}\tilde{\mathbf{D}}^{-\frac{1}{2}}\mathbf{H}^{(l)}\mathbf{W}^{(l+1)}),
\end{split}
\end{equation}
where $\mathbf{D}$ is the degree matrix, $\mathbf{H}^{(l)}$ is the node features for $l$-th layer, $\mathbf{W}$ is the GNN parameters, $\tilde{\mathbf{A}}=\mathbf{A}+\mathbf{I}$ is the adjacency matrix with self-loop. Then, GNN feeds the node features at the last layer to the NLL loss for training. 

\subsection{Structural Poisoning Attacks}  
%A major class of attacks against GNNs operates by poisoning the graph data. 
%In the real world, before training the GNN model, the analyst has to collect data from the Web platform to construct the graph $\mathcal{G}$. This gives the chance for the attacker to tamper with the data collection process, producing a \textit{poisoned graph} denoted as $\mathcal{G}^p$. 
We restrict our attention to \textit{structural poisoning attacks} \cite{mettack} since recent results \cite{gcnjaccard} show that modifying the topology is more harmful than modifying attributes. Alternatively, the semi-supervised nature of GNN is particularly suitable for the attacker to inject the poisons to the graph during the GNN training step~\cite{mettack}. Specifically, given a clean graph $\mathcal{G}$, the attacker modifies the structure (i.e., $\mathbf{A}$) of the graph by inserting/removing edges, resulting in a poisoned graph $\mathcal{G}^p = (\mathbf{V}_l, \mathbf{V}_u, \mathbf{X}, \mathbf{A}^p)$. Various attack methods are proposed to find the poisoned graph $\mathcal{G}^p$ such that when a GNN model is trained over $\mathcal{G}^p$, the prediction accuracy on the unlabeled nodes is minimized. The representative structural poisoning attacks are METTACK~\cite{mettack} and MinMax~\cite{topologyattack}, which formulates the attack as a discrete bi-level optimization problem. METTACK tackles this problem by greedily searching for the largest meta-gradient of all possible node pairs to ``flip" (insert or remove links) until reaches the budget. MinMax relaxes the discrete parameters to the continuous space and optimizes the problem via projection gradient descent~\cite{PGD} and obtain the poisoned graph by random sampling.

\section{Empirical Study of Sanitation}
In this section, we present an empirical study to illustrate the relationship between the poisoned graph sanitation performance and the robustness of GNN. In the meanwhile, we give a detailed description of the new metric--Effective Sanitation Ratio ($\mathsf{ESR}$) and utilize this metric to quantify the quality of the sanitized graph. 

\begin{wrapfigure}{r}{0.2\textwidth} 
    \includegraphics[width=0.2\textwidth]{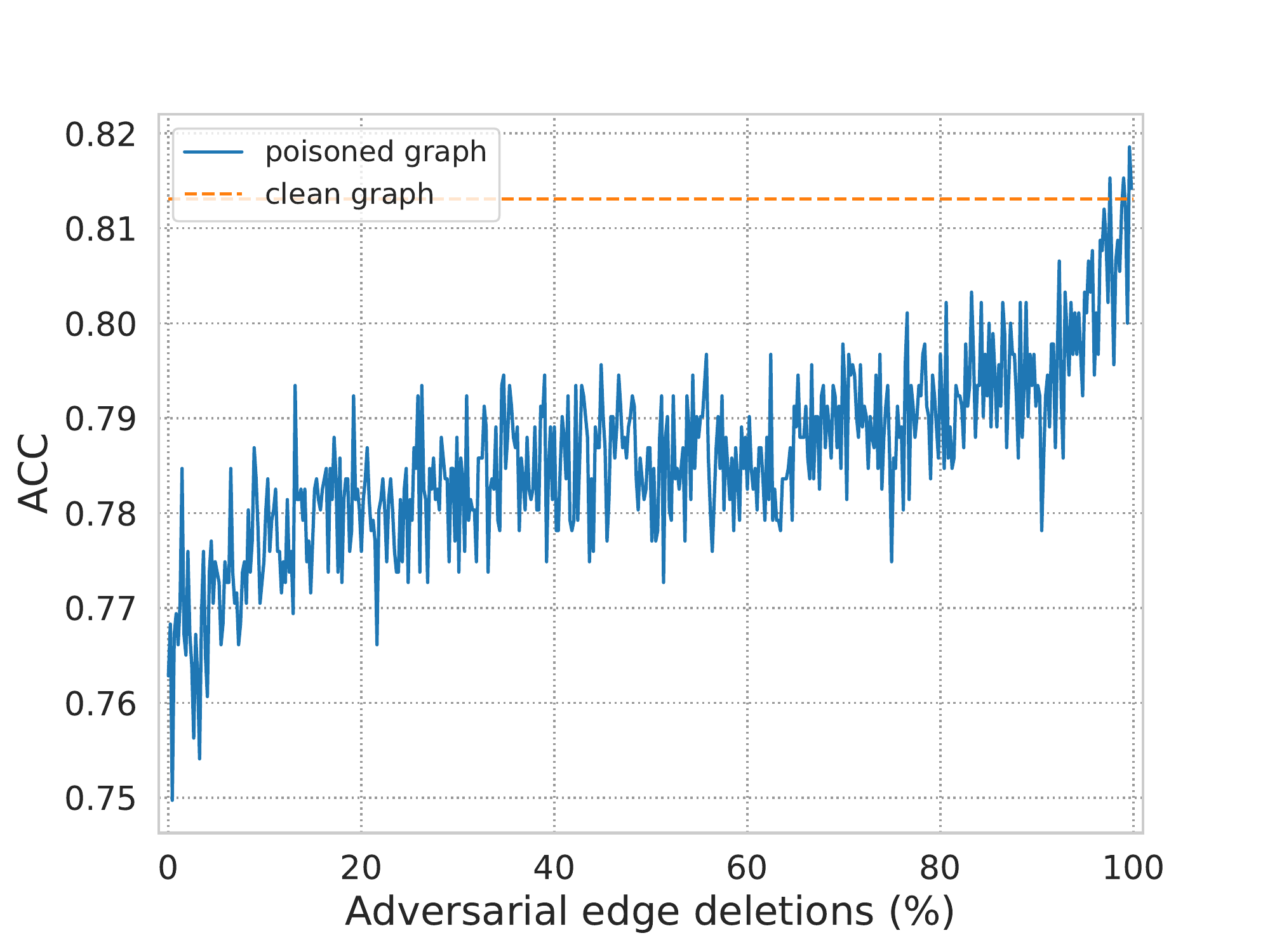}
    \caption{Pruning the adversarial links injected by METTACK with $10\%$ attacking power on Cora dataset.}
    \label{fig-adv_remove}
\end{wrapfigure}

\begin{figure}[h]
	\centering
	\begin{subfigure}[b]{0.236\textwidth}
		\centering
		\includegraphics[width=\textwidth,height=3.cm]{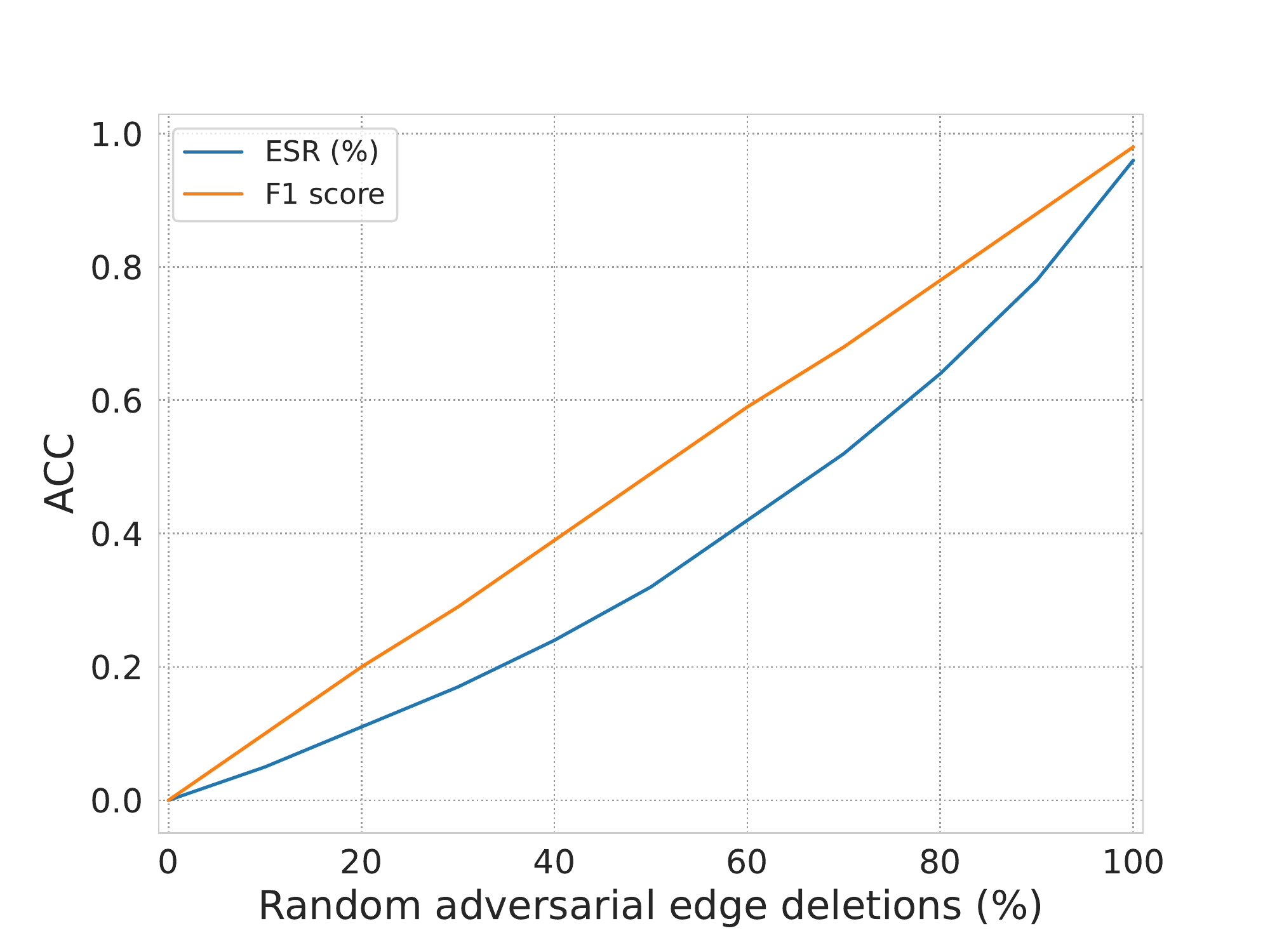}
		\caption{$\mathsf{ESR}$ and $F_{1}$ score}
	\end{subfigure}
	\hfill
	\begin{subfigure}[b]{0.236\textwidth}
		\centering
		\includegraphics[width=\textwidth,height=3.cm]{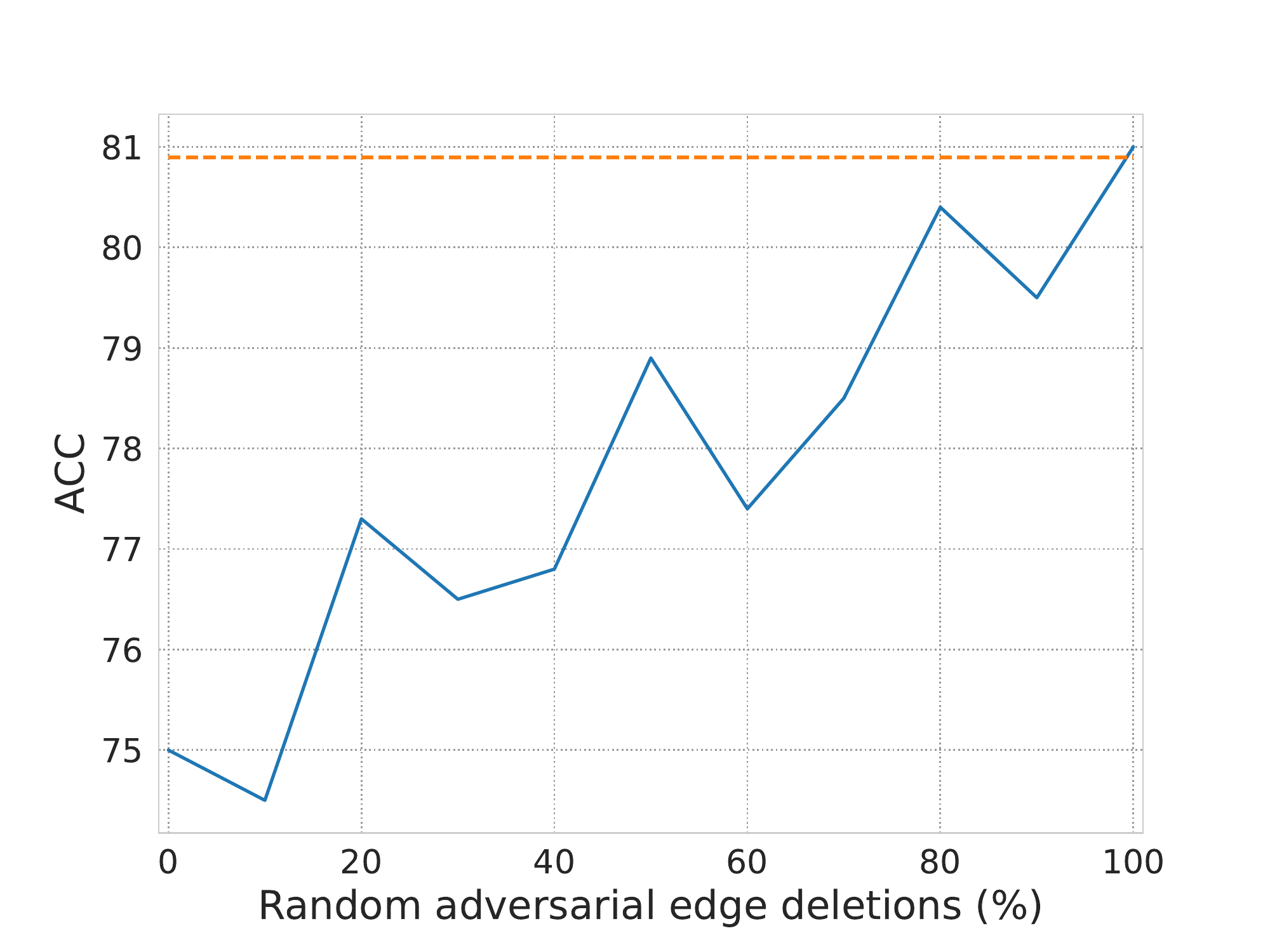}
		\caption{Classification accuracy}
	\end{subfigure}
	\caption{$\mathsf{ESR}$, $F_1$ score and mean node classification accuracy for randomly prune adversarial links and normal links.}
	\label{fig-random_remove}
\end{figure}

It is natural that pruning the injected adversarial links in the poisoned graph will drastically boost the node classification performance of GNN (METTACK and MinMax tend to insert links rather than delete links~\cite{gcnjaccard}). To verify this issue, we utilize METTACK to manipulate the Cora dataset with $10\%$ attacking power, i.e., the graph attacker can perturb the Cora dataset at most $506$ times. It is observed that the attacker injects $495$ adversarial links and deletes $11$ links. To validate the effectiveness of pruning the adversarial links, we delete the adversarial links in the poisoned graph one by one and report the mean testing accuracy over $5$ runs of GNN after each pruning. The trace plot of the mean accuracy for each pruning is depicted in Fig.~\ref{fig-adv_remove}. The orange line is the mean accuracy of the GNN trained on the clean graph. Fig.~\ref{fig-adv_remove} demonstrates that pruning more adversarial links in the poisoned graph will increase the performance of GNN until reaching the accuracy of the clean graph with slight deviations.
This phenomenon inspires us that under a fixed budget, a graph sanitizer's goal is to prune the adversarial links as much as possible to recover the poisoned graph to be similar to the clean graph. As a consequence, the sanitized graph will reach a more robust performance compared with the poisoned graph. To precisely quantify the number of adversarial links successfully pruned by the graph sanitizer, we design the metric $\mathsf{ESR}$ as:
\begin{equation}
\label{eqn-recover-ratio}
\mathsf{ESR}=\frac{|\mathcal{S}_{atk} \cap \mathcal{S}_{san}|}{|\mathcal{S}_{atk} \cup \mathcal{S}_{san}|}\in [0,1],
\end{equation}
where $\mathcal{S}_{san}$ is the pruned links set and $\mathcal{S}_{atk}$ is the adversarial links set. Naturally, a large $\mathsf{ESR}$ means most of the pruned links by the graph sanitizer are adversarial links. Moreover, we also introduce the $F_1$ score to quantify the sanitation performance as a reference to $\mathsf{ESR}$, i.e.,
\begin{equation}
\label{eqn-F1-score}
F_1=\frac{2|\mathcal{S}_{atk} \cap \mathcal{S}_{san}|}{|\mathcal{S}_{atk}|+|\mathcal{S}_{san}|}\in [0,1].
\end{equation}
To validate the effectiveness of $\mathsf{ESR}$ and $F_{1}$ score, we deploy another experiment on Cora dataset with random pruning. Under the $10\%$ attacking power, we restrict the attacker to randomly select $[0, 10\%, ..., 90\%, 100\%]$ of the attack budget to prune the true adversarial links while the remaining budget to prune the normal links. We then report the $\mathsf{ESR}$, $F_{1}$ score and the mean node classification accuracy over $5$ runs in Fig.~\ref{fig-random_remove}.
The experiment results demonstrate that higher $\mathsf{ESR}$ and $F_1$ score can lead to better node classification accuracy of GNN. Specially, when $\mathsf{ESR}$ and $F_1$ are $100\%$, the poisoned graph's accuracy is almost the same as clean graph. These phenomenons verify that a sanitized graph with good quality should have relatively high $\mathsf{ESR}$ and $F_1$ score. From this perspective, an excellent graph sanitizer aims at sanitizing the poisoned graph to obtain a sanitized graph with high $\mathsf{ESR}$ or $F_{1}$ values. 

%To this end, an excellent graph sanitizer aims at precisely spotting and removing the potential adversarial links as many as possible under a fixed budget. We report the sanitation performance of the preprocessing-based defense methods based on $\mathsf{ESR}$ and $F_1$ score in the experiments. 

\section{Problem of Graph Sanitation} 
Now, the task of graph data sanitization is clear: a defender  will try to identify the edges inserted/removed by the attacker from the poisoned graph $\mathcal{G}^p$. We denote $\Delta$ as the sanitation made to the adjacent matrix of $\mathcal{G}^p$ and assume that $|\Delta| \leq B$, meaning that the amount of sanitation (i.e., number of added and/or deleted edges) is bounded by $B$.
That is, through sanitization, the defender will obtain a sanitized graph denoted as $\mathcal{G}^R \triangleq \mathcal{G}^p \circledast \Delta$, here $\circledast$ denotes the sanitation operation at a high level. Then, the problem of graph sanitization can be formulated as the following optimization problem:
\begin{equation}
\label{eqn-origin-problem}
\Delta^* = \argmin_{\Delta} ||\mathcal{G} - \mathcal{G}^R||_d, \ \text{s.t.}\ \mathcal{G}^R \triangleq \mathcal{G}^p \circledast \Delta,\  |\Delta| \leq B,
\end{equation}
where $||\cdot||_d$ denotes a distance metric on the adjacent matrix at a high level. 
In words, the defender aims to clean the poisoned graph $\mathcal{G}^p$ so as to obtain a sanitized graph $\mathcal{G}^R$ that is similar to the original clean graph $\mathcal{G}$. We follow the semi-supervised node classification setting where the defender knows a subset of node labels. 
We note that $\mathcal{G}^p$ is the common knowledge available to both the attacker and the defender, while the ground-truth graph $\mathcal{G}$ is only available to the attacker.
%(the latter is typically assumed in poisoning attacks).
%{\color{green}(Liang: I'm a little confused after reading the formulation of graph sanitization. First, what's the difference between an arbitary modification (as long as the cost is within the budget $B$) and graph sanitization as shown in Eqn.~\ref{eqn-origin-problem}? Second, does $\circledast$ represents some kind of operators here?)}

\section{Poisoned Graph Sanitizer: $\mathsf{FocusedCleaner}$}

%\subsection{Our Ituitions}
\subsection{Importance of ``Focus" for Sanitation}
\label{sec-rethinking}
%Up to now, a surge of studies have drawn the conclusion that the graph attacker tends to add adversarial links to the clean graph \cite{gcnjaccard, survey}. Thus, there comes out a fundamental questions: \textit{Will it lead to a more clean sanitized graph if we restrict the preprocessing based methods to only delete links?} 
We explain our intuitions behind designing \textsf{FocusedCleaner} by discussing the limitations of the prior work GASOLINE \cite{GASOLINE} in sanitation. At a high level, instead of solving Problem~\eqref{eqn-origin-problem}, GASOLINE learns a new graph $\mathcal{G}'$ that could best \textit{restore the prediction accuracy} from the poisoned graph. 
%Then, $\mathcal{G}'$ is regarded as the recovered graph $\mathcal{G}^R$ (but not necessarily so!).
However, GASOLINE has its own bottlenecks. Our first observation is that previous studies \cite{gcnjaccard,survey} have demonstrated that in structural poisoning attacks the majority of manipulations are edge insertions rather than edge deletions.  This is a result of the optimization process to maximize attack performance. 
Thus, it is natural to restrict a sanitizer from ``flipping” (both inserting and deleting) edges to deleting edges only in a poisoned graph. Based on this, we consider a variation termed GASOLINE-D, which will only delete edges. 
%Without loss of generality, we use the previous work GASOLINE \cite{GASOLINE} as an example to verify this intuition. Specifically, we consider two variants of GASOLINE: GASOLINE-DT (the sanitizer can ``flip” links, DT is the shorthand of ``discrete topology” in \cite{GASOLINE}) and GASOLINE-Delete (the sanitizer can only delete links). In our experiment, we firstly use \textit{METTACK} to poison the Cora graph \cite{cora}. 
\begin{wraptable}{r}{5.5cm}
    \caption{Comparison of Sanitation.}
    \label{tab-DT-vs-Delete}
	\resizebox{0.6\columnwidth}{!}{%
		\begin{tabular}{l|rr}
			\toprule[1.pt]
			Methods &$\mathsf{ESR}$ (\%)&accuracy (\%)\\
			\hline
			GASOLINE     & $0.$  &$62.3$\\ 
			GASOLINE-D & $27.7$&$74.1$\\			
			\bottomrule[1.pt]
		\end{tabular}
	}
\end{wraptable} 
We then compare the performance of GASOLINE and GASOLINE-D in sanitizing the poisoned graph (Cora dataset) produced by METTACK. 
We note that METTACK is not restricted to inserting edges only.  
%Then we use  GASOLINE-DT and GASOLINE-Delete to clean the poisoned graph. 
We use the metric $\mathsf{ESR}$ (defined in Eqn.~\eqref{eqn-recover-ratio}) to measure the quality of sanitization, where a higher $\mathsf{ESR}$ means that more poisoned edges are pruned from the graph. Tab.~\ref{tab-DT-vs-Delete} shows the effective sanitation ratios as well as the node classification accuracy of GNN over the sanitized graph using two methods, respectively. 

%\begin{table}[H]
%	\centering
%	\caption{Comparison of Sanitation Results.}
%	\label{tab-DT-vs-Delete}
%	\resizebox{0.7\columnwidth}{!}{%
%		\begin{tabular}{c|cc}
%			\toprule[1.pt]
%			Methods &$\mathsf{ESR}$ (\%)&accuracy (\%)\\
%			\hline
%			GASOLINE     & $0.$  &$62.3$\\ 
%			GASOLINE-D & $27.7$&$74.1$\\			
%			\bottomrule[1.pt]
%		\end{tabular}
%	}
%\end{table}
There are a few crucial observations. First, GASOLINE also tends to insert edges (similar to attack) to best restore classification accuracy. This indeed shows that data sanitation is different from graph structural learning as GASOLINE has a zero recovery ratio. 
%Our crucial observation is that GASOLINE-DT also tends to insert edges into the graph to restore classification accuracy rather than removing edges. The possible reason is that inserting edges can more effectively affect GNNs' performance and GASOLINE basically works by inverting the attacking process of \textit{METTACK}. Thus, GASOLINE-DT will automatically learn to insert edges to try to \textit{restore classification accuracy}. In essence, GASOLINE-DT acts as more like a data augmentation method (improving model performance by modifying data) rather than a sanitization approach. In comparison, restricting from ``flipping” to only deleting edges indeed helps {GASOLINE-Delete} to sanitize poisoned graphs, thus leading to higher robustness performance.
%and the node classification accuracy of GCN in Tab.~\ref{tab-DT-vs-Delete}. At the same time, we also explore that GASOLINE-DT tend to add extra links to the poisoned graph to increase the node classification accuracy, that is, GASOLINE-DT is more like a data augmentation methods to the graph learning rather than a graph defender to mitigate the influence of the graph attacker. In Tab.~\ref{tab-DT-vs-Delete}, we find out that GASOLINE-DT fails to detect out the true adversarial links and thus leads to the degraded performance on node classification task. 
Second, GASOLINE-D outperforms GASOLINE in sanitation. Because by restricting to deleting edges only,  GASOLINE-D narrows the search space of the meta-gradients (which are used to determine which edges to be modified first) from $\mathcal{O}(N^{2})$ down to $\mathcal{O}(|E|)$. That is, we created a more ``focused" \textit{candidate space} for the sanitizer to identify maliciously modified edges. This inspired us to make the candidate space more focused to improve the sanitization performance. That is the ``focus" in \textsf{FocusedCleaner}.
%Guided by this intuition, we introduce a \textit{victim node detection} module to first identify those victim nodes whose connections are altered by attackers. Next, we further restrict the search space from $\mathcal{O}(|E|)$ to $\mathcal{O}(|E_{ano|})$, where $E_{ano}$ denotes the set of edges incident to the identified victim nodes.
In addition, the fact that GASOLINE-D outperforms GASOLINE in classification accuracy provides an interesting hypothesis: recovering the clean graph might be more effective in restoring model performance than structural learning. 

%Fortunately, the experiments show that restricting the meta-gradients search space of GASOLINE from $\mathcal{O}(N^{2})$ to $\mathcal{O}(E)$ will guide the defender to choose the candidate adversarial links by greedy search on the meta-gradients, as well as accurately sanitizing the poisoned graph will boost the defense performance of the GCN. Hence, another question comes out: \textit{Can we further help the defender to spot the candidate adversarial links more precisely and thus leading to more effective defense performance?} To tackle this issue, we implement the victim node detection to help the bi-level structural learner to decrease the search space from $\mathcal{O}(E)$ to $\mathcal{O}(E_{ano})$, where $E_{ano}$ is the number of predicted adversarial links. In the following context, we will detaily introduce our methods: \textsf{FocusedCleaner} and illustrate how it incorporates the GNN features and victim node detection results to pick out and prune the potential adversarial links. 

\subsection{Framework and Overview}
\label{sec-overview}
We are now ready to present \textsf{FocusedCleaner} (whose framework is shown in Fig.~\ref{fig-focusedcleaner-framework}) consisting of two modules: bi-level structural learning and victim node detection.
%; where the bi-level structural learning contains two components, i.e., GNN training, and poisoning edge selection. which is shown in Fig.~\ref{fig-framework}

We follow and improve the idea of structural learning (also used in GASOLINE) to solve the sanitation problem ~\eqref{eqn-origin-problem}, which is impossible to solve directly since the clean graph $\mathcal{G}$ is unknown to the defender. At a high level, we aim to find a better graph structure from which GNNs will make fewer erroneous predictions; however, we integrate a victim node detection module that will restrict the region of updates in the graph. Specifically, we reformulate the sanitation problem as the following (somehow high-level) bi-level optimization problem:
\begin{subequations}
\begin{align}
    %&\mathcal{G}^{R}=\argmin_{\mathcal{G}^{p}} \ \mathcal{L}_{S}(\mathbf{W}^{*}, \mathcal{G}^{p}, \mathcal{N}, \mathcal{V}, \mathcal{T}^{\prime}, \lambda_{\mathcal{V}}, \lambda_{\mathcal{T}^{\prime}}, \mathbf{Y}), \label{eqn-sanitation-outer}\\
    \mathcal{G}^{R}=&\argmin_{\mathcal{G}^{p}} \ \mathcal{L}_{S}(\mathbf{W}^{*}, \mathcal{G}^{p}, \mathcal{N}, \mathbf{Y}), \label{eqn-sanitation-outer} \\
    \text{s.t.} \quad 
    &\mathbf{W}^{*}=\mathcal{L}_{NLL}(\mathbf{W},\mathcal{G}^{p}, \mathbf{Y}), \label{eqn-sanitation-inner} \\
    %&\mathcal{A}, \ \mathcal{N}=VicNodeDec(\mathbf{W}^{*}, \mathcal{G}^{p}, 
    %\theta_{V}), \label{eqn-victim}\\
    &\mathcal{A}, \ \mathcal{N}=\mathsf{VicNodeDec}(\mathbf{W}^{*}, \mathcal{G}^{p}) \label{eqn-victim} \\
    &\|\mathcal{G}^{p}-\mathcal{G}^{p0}\|\leq B, \nonumber
\end{align}
\end{subequations}
where $\mathcal{L}_{S}$ and $\mathcal{L}_{NLL}$ are the losses for structure learning and training the GNN model, respectively. $\mathcal{G}^{p0}$ is the initially observed poisoned graph and we treat $\mathcal{G}^{p}$ as the optimization variables.
%$\mathcal{T}$, $\mathcal{V}$ and $\mathcal{T}^{\prime}$ are the training, validation and testing set,  $\mathcal{L}_{NLL}$ is the NLL loss of the GNN, 
Importantly, $\mathsf{VicNodeDec}(\cdot)$ denotes the victim node detection module 
%with the model parameters $\theta_{V}$ and 
that will predict the normal node set $\mathcal{N}$ and victim node set $\mathcal{A}$. Then $\mathcal{N}$ and  $\mathcal{A}$ will provide guidance to structure learning through two aspects. First, the victim node set $\mathcal{A}$ defines a candidate search region $\mathcal{E}^{p}$, which is the set of edges incident to \textit{at least one} victim node in $\mathcal{A}$. $\mathcal{E}^{p}$ is crucial when deciding which edges to clean. Specifically, we adopt a greedy strategy to solve the structure learning problem. That is, we compute the gradient of $\mathcal{L}_S$ with respect to each edge $e\in \mathcal{E}^{p}$ and pick to edge corresponding to the largest gradient to clean. Second, when calculating the loss $\mathcal{L}_S$ over the nodes, we make the restriction that those nodes should belong to the normal node set $\mathcal{N}$. 
%\mynote{In our ablation study (supplement)}, 
We show that this strategy (compared to also calculating losses over victim nodes) could improve sanitation performance in the ablation study. We note that the optimization problem omitted some details for easier illustration of the interaction between structure learning and victim node detection; the detailed construction is presented in Section~\ref{Sec-structure}.

%For Eqn.~\ref{eqn-sanitation-outer}, we utilize the greedy search (search for the edge with the largest gradient) to pick out the candidate edge to remove, i.e.,
%\begin{subequations}
%\begin{align}
%    &e(u,v)=\argmax_{u\in\mathcal{A}\lor v\in\mathcal{A}} \ \frac{\partial\mathcal{L}_{S}}{\partial\mathbf{A}^{p}_{uv}}, \\
%    &\mathcal{G}^{R}=\mathcal{G}^{p0}\setminus\mathcal{E}^{p}, \ \mathcal{E}^{p}=\{e(u,v)\}_{u,v\in\mathcal{A}},
%\end{align}
%\end{subequations}
%where $\mathcal{G}^{p0}$ is the initial poisoned graph. \mynote{Is $\mathcal{N}$ used? Is $\mathcal{E}^{p}$ correctly defined?}

Overall, the victim node detection module provides two kinds of ``focus" for the structural learning process. First, it restricts the loss to be calculated only over normal nodes.  Second, it provides a reduced search region for identifying malicious edges. In return, the structure learning will provide a steadily sanitized graph for victim node detection, resulting in more accurate identification of victim nodes.
%({\color{red} why and say that you have exp results to support this})
%Second, when updating the graph structure using greedy search (search for the edge with the largest gradient), it restricts a region for the update, where the region should be all the edges connecting to at least one victim node. Because, by definition, a victim node is a node whose incident edges are manipulated by attackers.

Next, we detail our realizations of the structural learning and victim node detection modules.

\subsection{Improved Bi-level Structure Learning Module}
\label{Sec-structure}
%{\color{red} Structure learning or structural learning, be consistent}
%{\color{red} Emphasize the difference in your design of the objective function}

We present the structure learning module without involving victim node detection for the moment. 
%Given a poisoned graph $\mathcal{G}^{p}$, structure learning aims to learn a refined graph structure $\mathbf{A}^{R}$ by pruning the adversarial noises. In detail, 
The typical bi-level structure learning  consists of two optimization processes: GNN training (inner optimization in Eqn.~\ref{eqn-sanitation-inner}) and poisoned edge selection (outer optimization in Eqn.~\ref{eqn-sanitation-outer}). 
%These two components reinforce each other during each sanitation step. 
We made a few improvements to the typical structure learning module to boost sanitation performance. 

First, we introduce an \textbf{attribute smoother} \cite{ProGNN} to the outer optimization, serving as a regularizer to preserve high-level graph homophily.  
%further improve the sanitation performance of this module by penalizing low level graph homophily, we introduce an attribute smoother \cite{ProGNN} to the outer optimization. 
Second, we also bring in an \textbf{adaptive testing loss} trick (detailed later) to the outer optimization process to augment more information thus better guiding the poisoned edge selector. Moreover, we only consider the normal unlabeled nodes for calculating the loss $\mathcal{L}_{S}$ to prevent the contaminated entropy caused by the victim nodes.
%At a high level, the attribute smoother {\color{red} which term is the attribute smoother? which term is the dynamic testing loss? do make readers guess.} penalizes low level graph homophily, thus can indirectly supervise the cleaner to assign higher meta-gradient on the potential adversarial links. In the meanwhile, the dynamic testing loss will introduce more information to the selector and boost the sanitation results. 
Specifically, we formulate the bi-level structure learning as follows:
%\mynote{check if equation 4 is correct}
\begin{subequations}
    \begin{align}
    \mathbf{A}^{R}=&\mathop{\arg\min}_{\mathbf{A}^{p}} \ \mathcal{L}_{S}(\mathbf{A}^{p},\mathbf{X},\mathbf{W}^{*},\mathbf{Y},\mathcal{N},\mathcal{V},\mathcal{T}^{\prime},\lambda_{\mathcal{V}},\lambda_{\mathcal{T}^{\prime}}), \nonumber\\
=& \mathop{\arg\min}_{\mathbf{A}^{p}} -\lambda_{\mathcal{V}}\sum_{i\in\mathcal{V}\cap\mathcal{N}}\sum_{c=1}^{C}\mathbf{y}_{ic}\ln \mathbf{S}_{ic}\nonumber\\ 
&-\lambda_{\mathcal{T}^{\prime}}\sum_{i\in\mathcal{T}^{\prime}\cap\mathcal{N}}\sum_{c=1}^{C}\hat{\mathbf{y}}_{ic}\ln \mathbf{S}_{ic} 
+\eta Tr(\mathbf{X}^{T}\mathbf{L}^{p}\mathbf{X}) \label{eqn-bilevel-outer}\\ 
\text{s.t.}\quad
        &\mathbf{W}^{*}=\mathop{\arg\min}_{\mathbf{W}} \ -\sum_{i\in\mathcal{T}}\sum_{c=1}^{C}\mathbf{y}_{ic}\ln \mathbf{S}_{ic}, \label{eqn-bilevel-inner}\\
		&\mathbf{S}=\text{softmax}(\hat{\mathbf{A}}^{p2}\mathbf{X}\mathbf{W}), \ \hat{\mathbf{y}}_{i}=\max(\mathbf{S}[I,:]), \label{eqn-bilevel-softmax}\\
		&\mathbf{L}^{p}=(\mathbf{D}^{p})^{-\frac{1}{2}}(\mathbf{D}^{p}-\mathbf{A}^{p})(\mathbf{D}^{p})^{-\frac{1}{2}}, \label{eqn-bilevel-smoother}\\
		&\frac{1}{2}\|\mathbf{A}^{p}-\mathbf{A}^{p0}\|_{1}\leq B, \ \lambda_{\mathcal{V}}+\lambda_{\mathcal{T}^{\prime}}=1, \label{eqn-bilevel-budget}
	\end{align}
\end{subequations}
%{\color{red} use , not . at the end of equation, if it is followed by "where"}
where $\tilde{\mathbf{A}}^{p}=\mathbf{A}^{p}+\mathbf{I}$, $\hat{\mathbf{A}}^{p}=(\tilde{\mathbf{D}}^{p})^{-\frac{1}{2}}\tilde{\mathbf{A}}^{p}(\tilde{\mathbf{D}}^{p})^{-\frac{1}{2}}$ is the normalized adjacency matrix, $\mathbf{A}^{R}$ is the sanitized graph, $\mathbf{A}^{p0}$ is the initial poisoned graph, $\mathbf{S}$ is the output of the two-layered linearized GNN \cite{sgc,mettack}, 
%{\color{red} the softmax of each node (softmax of a node, not formal)}, 
$Tr(\mathbf{X}^{T}\mathbf{L}^{p}\mathbf{X})$ is the attribute smoother. We note that $\mathbf{y}_{ic}$ and $\hat{\mathbf{y}}_{ic}$ denote the true label and the estimated label for nodes in the validation and test set, respectively.  $\lambda_{\mathcal{V}}$ and $\lambda_{\mathcal{T}^{\prime}}$ 
determine the relative importance of the validation loss and testing loss in the outer optimization (Eqn.~\eqref{eqn-bilevel-outer}). 
%and Eqn.~\ref{eqn-bilevel-inner} is the GNN inner training to update the GNN parameters $\mathbf{W}$.
Since the testing loss is highly skewed due to the adversarial noise in the early stage, we adaptively assign $\lambda_{\mathcal{V}}=1-\frac{t}{B}$ at $t$-th sanitation step. As a result, we can adaptively let the outer loss (Eqn.~\eqref{eqn-bilevel-outer}) to partially pay attention to the testing loss for each sanitation step in an ascending manner. The intuition is that at the beginning we only focus on the validation loss; as the poisoned graph is sanitized step by step, we pay more attention to the testing set to augment more information to guide the cleaner.

\subsection{Victim Node Detection Module}
We define a victim node as a node whose incident edges/non-edges have been manipulated by the attacker. 
%To identify the victim nodes in a poisoned graph, 
We design a victim node detection module based on two unsupervised methods termed \textbf{ClassDiv} and \textbf{LinkPred}, respectively.

\subsubsection{ClassDiv Detector}
%{\color{red} High-level ideas first, then implementation details}
%One common sense is to design an unsupervised learning method to pick out the victim nodes from the benign nodes. 
%One natural idea to detect victim nodes is by measuring how they are different from normal nodes.
%Thus, we refer to a series of metrics that quantify the class distribution disturbance among the target node's $2$-hop ego-network \cite{egonetwork} as features to train the unsupervised model.
Our basic idea is to identify some proper metrics that capture the difference between victim nodes and other normal nodes, and then use these metrics as input features to train an unsupervised model.

Intuitively, structural poisoning attacks against GNNs will break the \textit{homophily}~\cite{homophily} of the graph, i.e., similar nodes tend to be interconnected. As a result, a victim node would tend to be different from its neighbors since an attacker may add an edge to connect it with a very different node or delete an edge between it and a similar neighbor.
 %In light of this, we identify a set of metrics to quantify the class distribution (output of the pre-trained GNN) disturbance among the target node's $2$-hop ego-network \cite{egonetwork} and treat them as features for a node. %(what is "class distribution disturbance"? it is an unsupervised method, why there is a notion of "class" here?) 
In light of this, we choose three metrics $prox_{1}$, $prox_{2}$ \cite{prox12} and $Div_{JS}$ \cite{jsdiv} to measure how different a node is from its neighbors without loss of generality. Specifically, the three metrics utilize the \textit{logits} of nodes, denoted as $\mathbf{S}=\{s_{i}\}_{i=1}^{N}$, given by an inner trained GNN model in Eqn.~\eqref{eqn-bilevel-inner} to compute the measure of difference. 
%To build an effective and efficient victim node detection based on these three metrics, 

We build an unsupervised Deep Gaussian Mixture Model~\cite{DGMM} (DGMM) based on
%the engineering features derived from the class divergence metrics. 
carefully designed features. First, we define the soft class probability~\cite{KD} as 
\begin{equation}
    \begin{split}
        \mathbf{s}^{temp}_{i}=\frac{exp(\mathbf{Z}_{i}/T)}{\sum_{j}\mathbf{Z}_{j}/T}, \ \text{where} \ \mathbf{Z}_{i}=\hat{\mathbf{A}}^{p2}\mathbf{X}\mathbf{W}^{*}[i],
    \end{split}
\end{equation}
where $T$ is the temperature to be tuned. We denote $\mathbf{S}^{temp}=\{s_{i}^{temp}\}_{i=1}^{N}$. Adjusting $T$ will lead to different scales of class information and amplify the class divergence between a victim node and its neighbors or decrease the class divergence between a normal node and its neighbors. Second, we increase the efficiency of computing these metrics by \textbf{vectorization} technique. Specifically, we first define the pairwise KL divergence~\cite{KL}:
\begin{df}
	Let $\mathbf{S}=\{s_{i}\}_{i=1}^{N}$ be the output of the pre-trained GNN, the pairwise softmax KL divergence is defined as:
	\begin{subequations}
		\begin{align}
			&\quad Div_{KL}^{pair}(\mathbf{S})=[s_{i}\log(\frac{s_{i}}{s_{-i}})]_{i=1}^{N}\\
			&=[(s_{i}\log s_{i}-s_{i}\log s_{-i})]_{i=1}^{N}\\
			&=\begin{matrix}	\underbrace{[|(\mathbf{S}\odot\log\mathbf{S})\frac{\mathbf{1}^{T}}{N}|...|(\mathbf{S}\odot\log\mathbf{S})\frac{\mathbf{1}^{T}}{N}]} \\ N \ \text{times}
			\end{matrix}-\mathbf{S}\log\mathbf{S}^{T}.			
		\end{align}
	\end{subequations}
\end{df}
Here $\odot$ represents the element-wise product of two matrices. We thus can reformulate $prox_{1}$ and $prox_{2}$ by the defined pairwise softmax KL divergence:
\begin{df}
    Let $Div_{KL}^{pair}(\mathbf{S})$ be the pairwise KL divergence based on the output of GNN, $prox_{1}$ and $prox_{2}$ are formulated as:
    \begin{align}
	prox_{1}&=\frac{1}{\mathbf{D}^{p}}\odot[(Div_{KL}^{pair}(\mathbf{S})\odot \mathbf{A}^{p})\frac{\mathbf{1}^{T}}{N}],\\
	prox_{2}&=(\frac{1}{\mathbf{D}^{p}\odot(\mathbf{D}^{p}-\mathbf{I})})\odot([\mathbf{A}^{p}\odot(\mathbf{A}^{p}Div^{pair}_{KL}(\mathbf{S}))]\frac{\mathbf{1}^{T}}{N}).
\end{align}
\end{df}

\begin{proof}\renewcommand{\qedsymbol}{}
\begin{align*}
    prox_{1}(i)&=\frac{1}{\mathbf{D}^{p}_{i}}\sum_{j=1}^{N}\mathbf{A}^{p}_{ij}Div_{KL}^{pair}(\mathbf{S}_{ij})\\
    &=\frac{1}{\mathbf{D}^{p}_{i}}\sum_{j=1}^{N}(\mathbf{A}^{p}\odot Div_{KL}^{pair}(\mathbf{S}))[i,j],\\
    \text{so}, \ prox_{1}&=\frac{1}{\mathbf{D}^{p}}\odot[(Div_{KL}^{pair}(\mathbf{S})\odot \mathbf{A}^{p})\frac{\mathbf{1}^{T}}{N}].\\
    prox_{2}(i)&=\frac{1}{\mathbf{D}^{p}_{i}(\mathbf{D}^{p}_{i}-1)}\sum_{j=1}^{N}\sum_{k=1}^{N}\mathbf{A}^{p}_{ij}\mathbf{A}^{p}_{ik}Div^{pair}_{KL}(\mathbf{S}_{jk})\\
    &=\frac{1}{\mathbf{D}^{p}_{i}(\mathbf{D}^{p}_{i}-1)}\sum_{j=1}^{N}(\mathbf{A}^{p}_{ij}\sum_{k=1}^{N}(\mathbf{A}^{p}_{ik}Div^{pair}_{KL}(\mathbf{S}))[k,j])\\
    &=\frac{1}{\mathbf{D}^{p}_{i}(\mathbf{D}^{p}_{i}-1)}\sum_{j=1}^{N}(\mathbf{A}^{p}\odot(\mathbf{A}^{p}Div^{pair}_{KL}(\mathbf{S})))[i,j],\\
    \text{so}, \  prox_{2}&=(\frac{1}{\mathbf{D}\odot(\mathbf{D}-\mathbf{I})})\odot([\mathbf{A}^{p}\odot(\mathbf{A}^{p}Div^{pair}_{KL}(\mathbf{S}))]\frac{\mathbf{1}^{T}}{N}).
\end{align*}
\end{proof}

%(Sec.~C of supplement). 
%We leave the formulation of vectorized $prox_{1}$ and $prox_{2}$ in the supplement due to space limit. 
Third, we construct the input features for DGMM as:
\begin{subequations}
    \label{eqn-hybrid-features}
    \begin{align}
        &\tilde{\mathbf{X}}^{PCA}=\text{softmax}(\text{PCA}(\mathbf{X}, C)/T), \\
        &P_{1}^{X}=prox_{1}(\tilde{\mathbf{X}}^{PCA}), P_{2}^{X}=prox_{2}(\tilde{\mathbf{X}}^{PCA}), \\
        &P_{1}^{S}=prox_{1}(\mathbf{S}^{temp}), P_{2}^{S}=prox_{2}(\mathbf{S}^{temp}), \\
        &D^{X}=Div_{JS}(\tilde{\mathbf{X}}^{PCA}), D^{S}=Div_{JS}(\mathbf{S}^{temp}),\\
        &\mathbf{M}=(P_{1}^{X}, P_{2}^{X}, D^{X}, P_{1}^{S}, P_{2}^{S}, D^{S}),
    \end{align}
\end{subequations}
where $\mathbf{M}$ is the input features that distill the class divergence information for each node. 
%$\text{PCA}(\cdot)$ is the traditional feature dimension algorithm~\cite{PCA}. 
We use PCA~\cite{PCA} to reduce the attributes dimension to the class number $C$. It is worth noting that if the graph does not have attributes, we only consider $\mathbf{M}=(P_{1}^{S}, P_{2}^{S}, D^{S})$. Next, we feed the hybrid features $\mathbf{M}$ into the unsupervised DGMM~\cite{DGMM} to identify the victim nodes as follows:
\begin{subequations}
    \begin{align}
	&\mathbf{H}_{1}=\text{ReLU}(\mathbf{w}_{1}\mathbf{M}+\mathbf{b}_{1}), \\
        &\hat{\gamma}=\text{softmax}(\mathbf{w}_{2}\mathbf{H}_{1}+\mathbf{b}_{2}),\\
	&\hat{\phi}_{k}=\sum_{i=1}^{N}\frac{\hat{\gamma}_{ik}}{N}, \ \hat{\mu}_{k}=\frac{\sum_{i=1}^{N}\hat{\gamma}_{ik}\mathbf{M}_{i}}{\sum_{i=1}^{N}\hat{\gamma}_{ik}},\\
    &\hat{\mathbf{\Sigma}}_{k}=\frac{\sum_{i=1}^{N}\hat{\gamma}_{ik}(\mathbf{M}_{i}-\hat{\mu}_{k})(\mathbf{M}_{i}-\hat{\mu}_{k})^{T}}{\sum_{i=1}^{N}\hat{\gamma}_{ik}},\\
    &E(\mathbf{M})=-\log(\sum_{k=1}^{K}\hat{\phi}_{k}\frac{\exp(-\frac{1}{2}(\mathbf{M}_{i}-\hat{\mu}_{k})^{T}\hat{\mathbf{\Sigma}}_{k}^{-1}(\mathbf{M}_{i}-\hat{\mu}_{k}))}{\sqrt{|2\pi \hat{\mathbf{\Sigma}}_{k}|}}),
    \end{align}
\end{subequations} 
where $\hat{\gamma}_{ik}$ denotes the probability of the node $v_{i}$ belonging to the cluster $k$ (we set $K$ equal to the class number $C$), $E(\mathbf{M})$ is the energy function of DGMM. We adopt the Adam optimizer \cite{adam} to optimize the energy function $E(\mathbf{M})$. 
After training, we compute $E(\mathbf{M}_{i})$ for each node and higher $E(\mathbf{M}_{i})$ tend to be anomalous. For evaluation, we denote the $\tau$-th quantile of the energy scores as $\alpha=Q_{\tau}(E(\mathbf{M}))$ and regard a fraction of $(1-\tau)$ of the nodes as the victims. However, setting a fixed value of $\tau$ is not a wise choice since, as the graph is sanitized, there should be fewer victim nodes. To address this, we use an \textbf{adaptive threshold} for victim node detection. Inspired by the \textbf{momentum trick}~\cite{momentum}, we update the threshold at each sanitation step for victim node detection as:
\begin{subequations}
    \begin{align}
	&\kappa_{t}=\beta\alpha_{t}+(1-\beta)\kappa_{t-1}, \\ 
    &\text{where} \ \ \kappa_{0}=Q_{\tau}^{(0)}(E(\mathbf{M})), \alpha_{t}=Q_{\tau}^{(t)}(E(\mathbf{M})),
    \end{align}
\end{subequations}
where $\kappa_{t}$ is the threshold for the $t$-th sanitation step, $\alpha_{t}$ is the $\tau$-th quantile based on the energy score at $t$-th sanitation step, $\beta$ is the hyperparameter to balance the quantile at $t$-th step and the threshold at last step. Thereafter, we adaptively predict the victim node set at $t$-th sanitation step as:
\begin{align}
\mathcal{A}_{t}=\{i\in\{1,...,N\}|E(\mathbf{M}_{i})>\kappa_{t}\}, \  \forall t\in\{1,...,B\}. 
\end{align}
The remaining nodes are regarded as normal.
%nodes and belong to the normal node set $\mathcal{N}$.
%\begin{figure}[h]
 %   \centering
%	\includegraphics[width=0.4\textwidth,height=3.5cm]{figures/DGMM dist.png}
%	\caption{Histogram of the energy scores for the DGMM. Red curve is the density of the poisoned graph, orange curve is the density of the sanitized graph.}
%	\label{fig-DGMM-plot}
%\end{figure}

\subsubsection{LinkPred Detector} 
%{\color{red} Again, high-level ideas first, then implementation details}
Alternatively, we also observe that link prediction~\cite{LinkPred} can be used to detect the potential victim nodes. Intuitively, a link prediction model can assign each link a probability, and a link with a lower probability tends to be maliciously injected. We then regard a node as a victim if it is incident to a malicious link. To this end, we utilize the node embeddings output from the inner training (Eqn.~\eqref{eqn-bilevel-inner}) of GNN as input and build up a two-layer MLPs \cite{mlp} with an inner-product layer as the prediction model (termed as \textbf{LinkPred}):
\begin{subequations}
	\begin{align}
		&\mathbf{H}_{1}=\text{ReLU}(\mathbf{w}_{1}\text{concat}(\mathbf{Z}| \tilde{\mathbf{X}}^{PCA})+\mathbf{b}_{1}),\\
		&\mathbf{H}_{2}=\mathbf{w}_{2}\mathbf{H}_{1}+\mathbf{b}_{2}, \ \mathbf{H}_{3}=\text{Sigmoid}(\mathbf{H}_{2}\cdot\mathbf{H}_{2}^{T}).
	\end{align}
\end{subequations} 
Similarly, we remove $\tilde{\mathbf{X}}^{PCA}$ if the graph does not have attributes. The loss function is the reweighting binary cross-entropy loss with the reweighting parameter $\gamma=\frac{N^{2}-|E|}{|E|}$ to tackle the imbalanced problem when training. We choose the threshold $\tau_{lp}$ which can max the G-mean~\cite{Gmean} of the prediction results. We then obtain the victim node set at $t$-th sanitation step as:
\begin{equation}
    \begin{split}
        \mathcal{A}_{t}=\{i\in\{1,...,N\}|\forall (i,j)\in \mathcal{E}, \mathbf{H}_{3}(i,j)<\tau_{lp}^{t}\},
    \end{split}
\end{equation}
where $\mathcal{E}$ is the link set, $\tau_{lp}^{t}$ is the threshold at $t$-th step. Finally, this victim node detection module then interacts with structure learning as illustrated in Section~\ref{sec-overview} to perform sanitation. The algorithm of $\mathsf{FocusedCleaner}$ is shown in Alg.~\ref{alg-focusedcleaner}.
\begin{algorithm}[h]
	\caption{\textsf{FocusedCleaner}}
	\label{alg-focusedcleaner}
	\textbf{Input}: Poisoned graph $\mathcal{G}^{p0}=\{\mathbf{A}^{p0},\mathbf{X}\}$, graph parameters $\mathbf{A}^{p}$, link set $\mathcal{E}^{p}$ for the poisoned graph $\mathcal{G}^{p0}$, sanitation budget $B$, hyperparameters for ClassDiv based victim node detection: $T$, $\beta$, and $\tau$, feature smoothing penalizer $\eta$, training dataset $\mathcal{T}$, validation dataset $\mathcal{V}$ and testing dataset $\mathcal{T}^{\prime}$, training node labels $\mathbf{y}_{\mathcal{T}}$ and validation node labels $\mathbf{y}_{\mathcal{V}}$.\\
	\textbf{Output}: Sanitized graph $\mathcal{G}^{R}=\{\mathbf{A}^{R},\mathbf{X}\}$.\\
	\begin{algorithmic}[1] %[1] enables line numbers
		\STATE Let $t=0$, initialize parameters $\mathbf{A}^{p}=\mathbf{A}^{p0}$.
		\STATE $\tilde{\mathbf{X}}^{PCA}=\text{softmax}(PCA(\mathbf{X})/T)$.
		\WHILE{$t\leq B$}
        \STATE Inner training $\mathbf{W}^{*}=\mathcal{L}_{NLL}(\mathbf{W},\mathbf{A}^{p},\mathbf{X},\mathbf{Y})$.
		\IF{ClassDiv-based victim node detection}
		\STATE $\mathbf{S}^{temp}, \hat{\mathbf{y}} \leftarrow \text{GNN}(\mathbf{W}^{*},\mathbf{A}^{p},\mathbf{X}$).\\
		\STATE $\hat{\mathbf{y}}_{ano}=\text{DGMM}_{\theta^{*}}(\mathbf{A}^{p},\mathbf{X},\tilde{\mathbf{X}}^{PCA},\mathbf{S}^{temp}, \beta, \tau).$\\
		\ELSIF{LinkPred-based victim node detection}
		\STATE $\mathbf{Z}, \hat{\mathbf{y}} \leftarrow \text{GNN}(\mathbf{W}^{*},\mathbf{A}^{p},\mathbf{X}$).\\
		\STATE $\hat{\mathbf{y}}_{ano}=\text{LinkPred}_{\theta^{*}}(\mathbf{A}^{R},\mathbf{Z}, \tilde{\mathbf{X}}^{PCA}).$
		\ENDIF
		\STATE Normal node set $\mathcal{N}=\{v\in\mathcal{G}^{p0}|\hat{\mathbf{y}}_{ano}[v]=0\}.$\\
		\STATE Set $\lambda_{\mathcal{V}}=1-\frac{t}{B}$ and obtain the meta-gradient $\frac{\partial\mathcal{L}_{S}(\mathbf{W}^{*}, \mathbf{A}^{p},\mathbf{X},\mathcal{N},\mathbf{Y})}{\partial\mathbf{A}^{p}}$.
		\STATE Set $\frac{\partial\mathcal{L}_{S}(\mathbf{W}^{*}, \mathbf{A}^{p},\mathbf{X},\mathcal{N},\mathbf{Y})}{\partial\mathbf{A}^{p}}[u,v]=0$, $\forall u,v\in\mathcal{N}$.\\
		\STATE $(u*,v*)=\argmax_{\{(u,v)\in \mathcal{E}^{p}\}}\ \frac{\partial\mathcal{L}_{S}(\mathbf{W}^{*}, \mathbf{A}^{p},\mathbf{X},\mathcal{N},\mathbf{Y})}{\partial\mathbf{A}^{p}}$\\
		\STATE $\mathbf{A}^{p}\leftarrow\mathbf{A}^{p}\setminus \{u*,v*\}$\\
		\ENDWHILE
	\end{algorithmic}
\end{algorithm} 

\subsection{Time and Space Complexity}
The time complexity of $\textsf{FocusedCleaner}_{CLD}$ involves the computation of the meta-gradients for all the node pairs and the hybrid features $\mathbf{M}$ in Eqn.~\ref{eqn-hybrid-features}. By utilizing the vectorization technique for computing $prox_{1}$ and $prox_{2}$, the time complexity of computing $\mathbf{M}$ decreases from $\mathcal{O}(N^{2})$ to $\mathcal{O}(N)$. While it is known that the time complexity of GASOLINE~\cite{GASOLINE} is $\mathcal{O}(BN^{2}d^{2})$, where $d$ is the average nodes degree. Since $\textsf{FocusedCleaner}_{CLD}$ narrow down the search space from $\mathcal{O}(N^{2})$ to $\mathcal{O}(|E_{ano}|)$ ($E_{ano}$ is the adversarial links set), the time complexity of structural learning module is $\mathcal{O}(B|E_{ano}|d^{2})$. After augmenting the victim node detection module, the time complexity of $\textsf{FocusedCleaner}_{CLD}$ is $\mathcal{O}(B|E_{ano}|d^{2}N)$. Similar to GASOLINE, the space complexity of $\textsf{FocusedCleaner}_{CLD}$ is $\mathcal{O}(|E|)$ if the graph is stored as sparse matrix and $\mathcal{O}(N^{2})$ for dense matrix.

\section{Experiments}
%\mynote{The fonts of all specific terms, like Mettack, other GNN models, should be consistent}
In this section, we evaluate the sanitation performance of \textsf{FocusedCleaner} and how it contributes to the robustness of GNN-based node classification and answering the following three vital questions:
\begin{itemize}
    \item How is the performance of \textsf{FocusedCleaner} in sanitizing poisoned graphs?
    \item Is it necessary to augment the bi-level structural learning framework with victim node detection?
    \item How can \textsf{FocusedCleaner} contribute to the adversarial robustness of GNNs for node classification?
\end{itemize}

\subsection{Experimental Settings}
\subsubsection{Datasets}
We evaluate the methods over three standard datasets: Cora~\cite{citationgraph, sen2008collective}, Citeseer, and Polblogs, where Polblogs has no node features. Since Polblogs does not have node attributes, we use the identity matrix to represent its attribute matrix. We follow the same setting as GASOLINE~\cite{GASOLINE} and randomly split the datasets into training ($10\%$), validation($10\%$) and testing ($80\%$) dataset. The statistics of the datasets are presented in Tab.~\ref{tab-dataset}. We only consider the largest connected component (LCC)~\cite{mettack} of each graph.
\begin{table}[h]
	\centering
	\caption{Dataset statistics.}
	\label{tab-dataset}
	\resizebox{0.8\columnwidth}{!}{%
		\begin{tabular}{l|rrrr}
			\toprule[1.pt]
			Datasets &$N_{LCC}$&$E_{LCC}$&Classes&Features\\
			\hline
			Cora     & $2485$  & $5069$  & $7$   & $1433$\\ 
			Citeseer & $2110$  & $3668$  & $6$   & $3703$\\
			Polblogs & $1222$  & $16714$ & $2$   &   /   \\
			\bottomrule[1.pt]
		\end{tabular}
	}
\end{table}

%{\color{red} Explicitly list the baseline methods here (not in the appendix). It is more eye-catching. Divide them into two types}
%We conduct experiments on three benchmark datasets: Cora \cite{cora}, Citeseer \cite{cora} and Polblogs \cite{polblogs}.
GCN~\cite{gcn} is used as the default GNN model for all the evaluations. METTACK and MinMax are employed as the two representative global structural attacks to test all the sanitation and defense methods. 
%For more settings please refer to Sec.~B of the supplement.

\subsubsection{Baseline Methods}
To evaluate the sanitation performance of \textsf{FocusedCleaner}, we compare it with other state-of-the-art graph structure learning methods (which could also be used as sanitizers): \textbf{GCN-Jaccard}, \textbf{maskGVAE}, and \textbf{GASOLINE-D} (we ignore GASOLINE here since GASOLINE has been proved to be less effective than GASOLINE-D in Sec.~6.1). For a fair comparison, we restrict all these sanitizers to deleting edges only. We note that since these sanitizers can clean the poisoned data, they can act as \textit{preprocessing-based} defense methods. Meanwhile, there are other robust GNN models that are designed to defend against attacks; we term them as \textit{robust-model-based} defense methods. Thus, to evaluate the defense performance of \textsf{FocusedCleaner}, we compare it with both preprocessing-based and notable robust-model-based methods, including \textbf{ProGNN} \cite{ProGNN}, \textbf{RGCN} \cite{RGCN}, \textbf{MedianGNN} \cite{mediangnn}, \textbf{SimPGCN}, \textbf{GNNGUARD} \cite{simpgcn}, \textbf{ElasticGNN} \cite{elasticgnn} and \textbf{AirGNN} \cite{airgnn}. We use \textsf{Deeprobust} \cite{deeprobust} to implement two typical structural poisoning attack methods: METTACK and MinMax and two defense methods \textbf{GNN-Jaccard}~\cite{gcnjaccard} and \textbf{ProGNN}~\cite{ProGNN}. We use \textsf{GreatX} \cite{greatx} to implement RGCN, MedianGNN, SimPGCN, GNNGUARD, ElasticGNN and AirGNN. We implement GASOLINE-D and maskGVAE using the codes provided by the authors. The descriptions of the preprocessing-based models are listed below:
\begin{itemize}
	\item \textbf{GCN-Jaccard} It prunes links based on the nodes' attribute similarities.
	\item \textbf{maskGVAE} It trains the graph partition task and adopts the clustering results to supervise the graph autoencoder.
	\item \textbf{GASOLINE-D} It adopts the bi-level optimization to augment the optimal graph structure by pruning links for the node classification task. It is a variant of GASOLINE.
\end{itemize}
While the description of robust GNNs are:
\begin{itemize}
	\item \textbf{RGCN} It utilizes gaussian distributions to represent node features and assigns an attention mechanism to penalize nodes with large variance.   
	\item \textbf{ProGNN} It jointly trains a dense adjacency matrix and the node classification with three penalties: feature smoothness, low-rank and sparsity.
	\item \textbf{MedianGNN} It utilizes the median aggregation layer to enhance the robustness of GNN.
	\item \textbf{SimPGCN} It utilizes a kNN graph to capture the node similarity and enhance the node representation of the GNN.
	\item \textbf{GNNGUARD} It utilizes the cosine similarity to calculate the link pruning probability during GNN training.
	\item \textbf{ElasticGNN} It introduces $L1$-norm to graph signal estimator and proposes elastic message passing during GNN training.
	\item \textbf{AirGNN} It adopts an adaptive message passing scheme to enhance GNN with adaptive residual.
\end{itemize}
The default hyperparameter settings for \textsf{FocusedCleaner} are: $\tau=0.6$, $\eta=10^{-4}$, $\beta=0.3$ and $T=2$. We tune $\tau$ from $\{0.5,0.6,0.7,0.8,0.9\}$, $\eta$ ranged from $\{0,10^{-6},10^{-5},10^{-4},10^{-3},10^{-2},10^{-1},1\}$, $\beta$ ranged from $\{0,0.1,0.2,0.3,0.4,0.5,0.6,0.7,0.8,0.9,1\}$ and $T$ ranged from $\{1,2,3,4,5\}$. Since the Polblogs does not have node attributes, we set $\eta=0$ and remove the corresponding features $\tilde{\mathbf{X}}^{PCA}$. For all the defense models, we report the
mean accuracy of $10$ runs with different random seeds. The dimension of the GCN hidden layer is $16$. %For all preprocessing-based methods, we assume they only delete links for fair comparison. 
%Since the defender does not know the attack power in advance, we set the sanitation ratio (is it defined) as $10\%$ for all the preprocessed-based methods based on the link numbers for the poisoned graph (For GCN-Jaccard, we are tuning the threshold as the number of deleting links is near $10\%$ for the poisoned graph with different attacking powers, i.e., the threshold for Cora, Citeseer and Polblogs are $6.3$, $8.15$ and $1.2$). 
For METTACK and MinMax, we consider $5$ different attacking powers: $5\%$, $10\%$, $15\%$, $20\%$ and $25\%$ based on the number of links for clean graph. 
%A more detailed description of these models and other experiment settings are provided in Sec.~B of the supplement.

\subsection{Analysis of Sanitation Performance}
%\subsubsection{Sanitation Results}
%To compare the pruning accuracy among different preprocessing based methods, we define the recovery ratio $\mathcal{R}$ as the \textbf{Jaccard Coefficient} between the adversarial links set and the pruning links set, i.e.,
\subsubsection{Evaluation Metric} 
Let $\mathcal{S}_{atk}$ and $\mathcal{S}_{san}$ be the set of malicious edges injected by the attacker and the set of suspected malicious edges produced by a sanitizer. We define the defender's \emph{Affordable Sanitation Budget} as the ratio $\mathcal{R}_{ASB} = |\mathcal{S}_{san}|/|E|$, where $|E|$ denotes the number of edges in the poisoned graph. Practically, $\mathcal{R}_{ASB}$ measures how much effort the defender could \emph{afford} to deal with the sanitation result $\mathcal{S}_{san}$. Because in practice, the defender usually needs to invest manpower to examine the found malicious edges for verification. Thus, setting a large $\mathcal{R}_{ASB}$ for a sanitizer would be overloaded. We follow the baseline method GASOLINE-D to set $\mathcal{R}_{ASB}$ a fixed value of $10\%$ throughout the evaluation (For GCN-Jaccard, we are tuning the threshold as the number of deleting links is near $10\%$ for the poisoned graph with different attacking powers, i.e., the threshold for Cora, Citeseer and Polblogs are $6.3$, $8.15$ and $1.2$); later on, we also discuss the effects of $\mathcal{R}_{ASB}$ on sanitation performance.

%We now define an \emph{Effective Sanitation Ratio} ($\mathsf{ESR}$) to measure the sanitation performance, as follows:
As previously mentioned in Sec.~4, we utilize two metrics $\mathsf{ESR}$ and $F_1$ score to quantify the sanitation performance of the graph sanitizer. We select $\mathsf{ESR}$ and $F_1$ score as the main metrics because, on the one hand, they measure how good the sanitizer is in successfully spotting malicious edges (i.e., $\mathcal{S}_{atk} \cap \mathcal{S}_{san}$) and on the other hand, they prevent a sanitizer from irresponsibility output too many suspected edges  (i.e., $\mathcal{S}_{atk} \cup \mathcal{S}_{san}$ and $|\mathcal{S}_{atk}|+|\mathcal{S}_{san}|$), thus causing lots of false alarms. Moreover, we define another metric \textit{Coverage Ratio}, i.e., 
\begin{align}
    \mathsf{CR}=\frac{|\mathcal{S}_{atk} \cap \mathcal{S}_{san}|}{|\mathcal{S}_{atk}|}
\end{align}
to measure the percentage of malicious edges that are successfully identified. 
%We also use the standard $F_1$ score for evaluation.
\begin{figure*}[htp]
	\centering
	\begin{subfigure}[b]{0.33\textwidth}
		\centering
		\includegraphics[width=\textwidth,height=3.5cm]{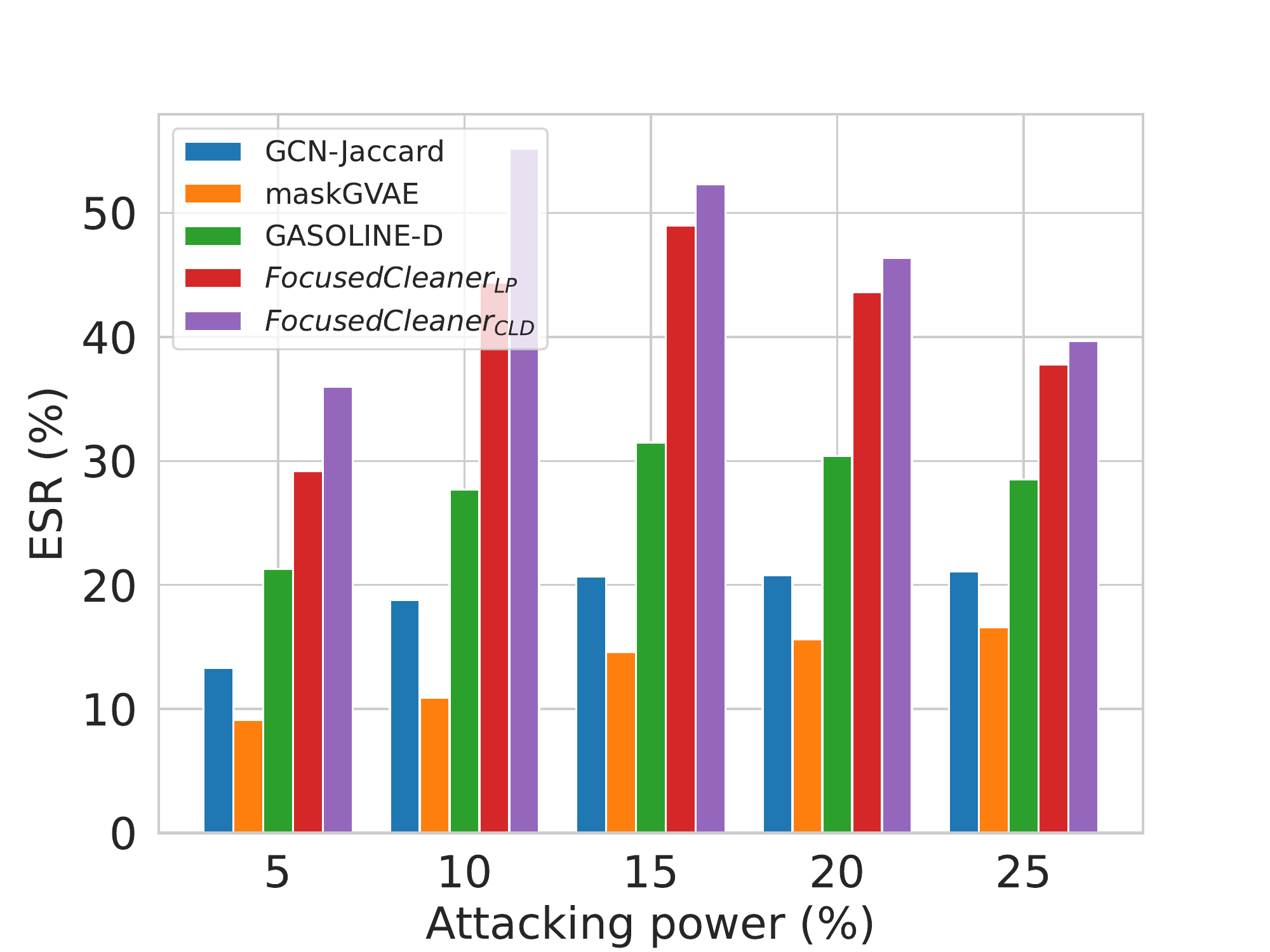}
		\caption{Cora for METTACK}
	\end{subfigure}
	\hfill
	\begin{subfigure}[b]{0.33\textwidth}
		\centering
		\includegraphics[width=\textwidth,height=3.5cm]{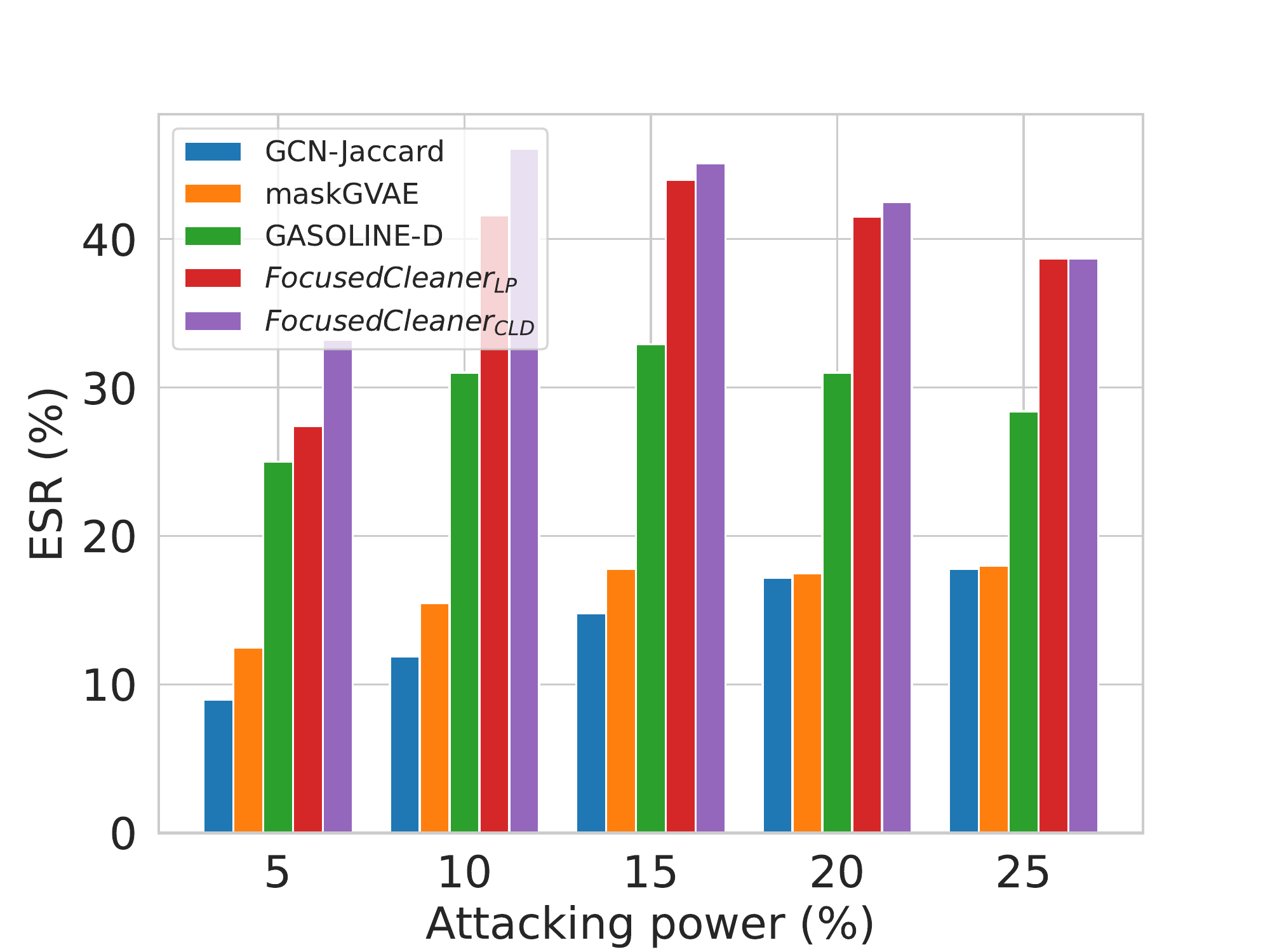}
		\caption{Citeseer for METTACK}
	\end{subfigure}
    \hfill
	\begin{subfigure}[b]{0.33\textwidth}
		\centering
		\includegraphics[width=\textwidth,height=3.5cm]{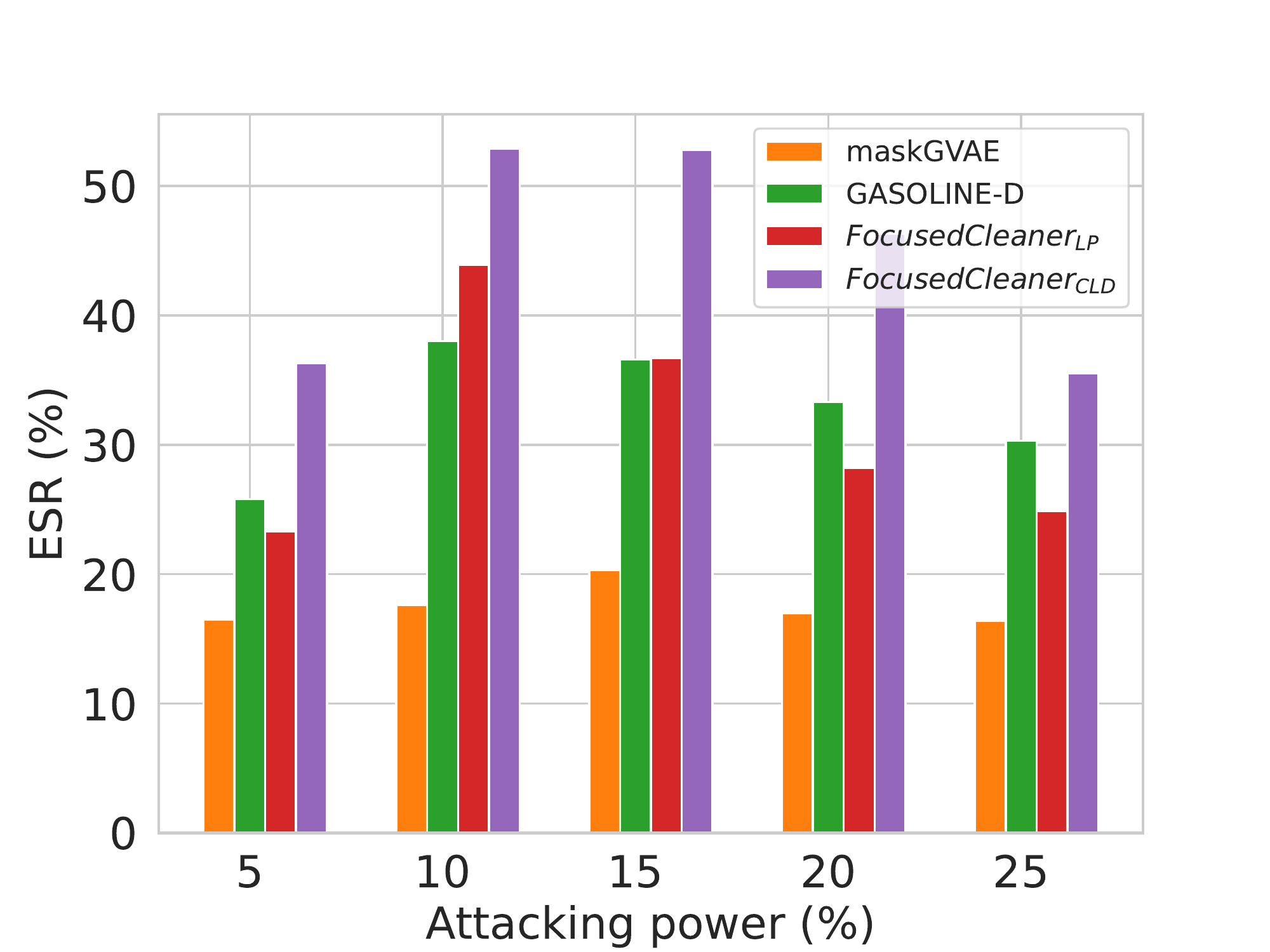}
		\caption{Polblogs for METTACK}
	\end{subfigure}
    \hfill
	\begin{subfigure}[b]{0.33\textwidth}
		\centering
		\includegraphics[width=\textwidth,height=3.5cm]{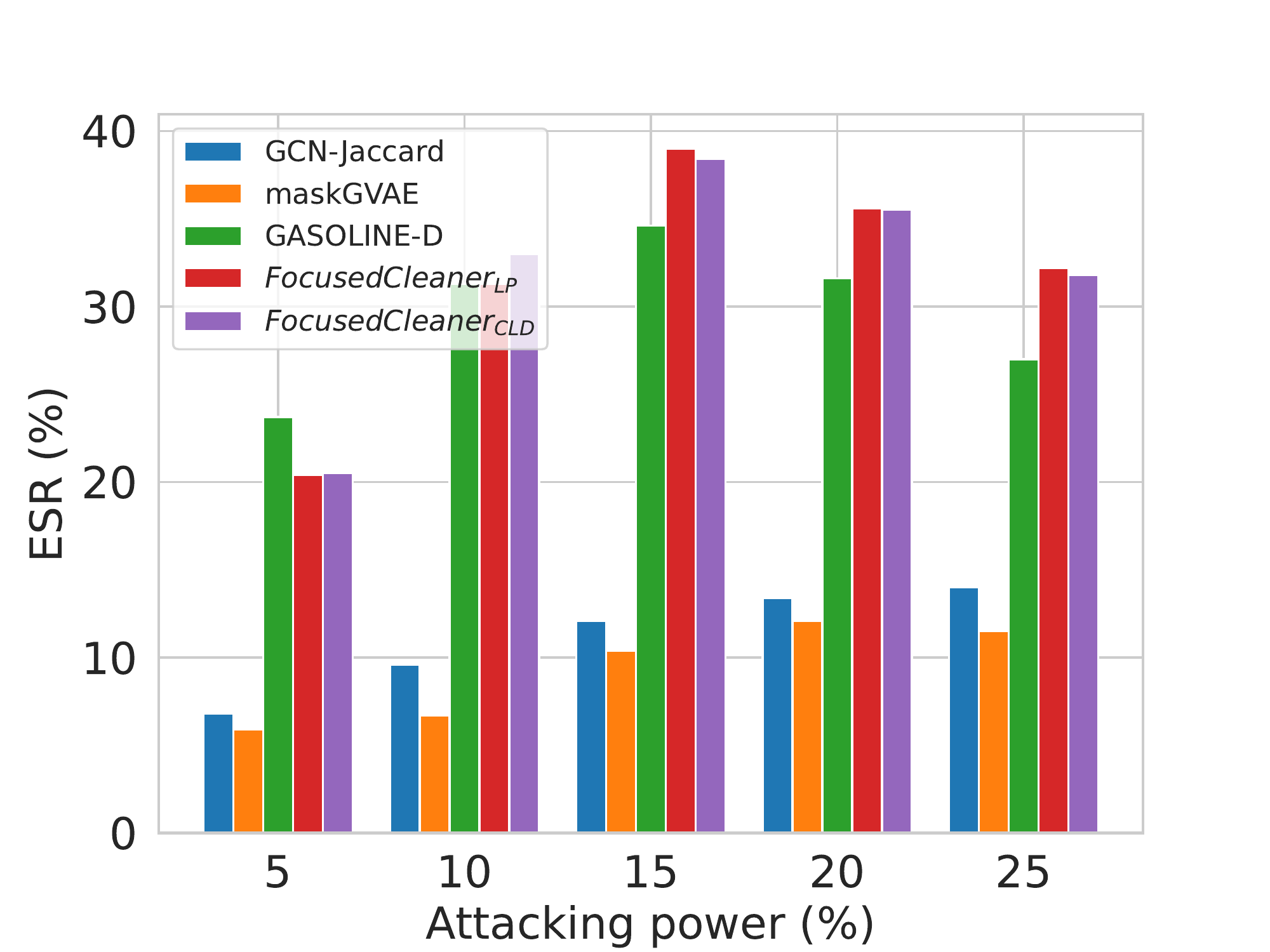}
		\caption{Cora for MinMax}
	\end{subfigure}
	\hfill
	\begin{subfigure}[b]{0.33\textwidth}
		\centering
		\includegraphics[width=\textwidth,height=3.5cm]{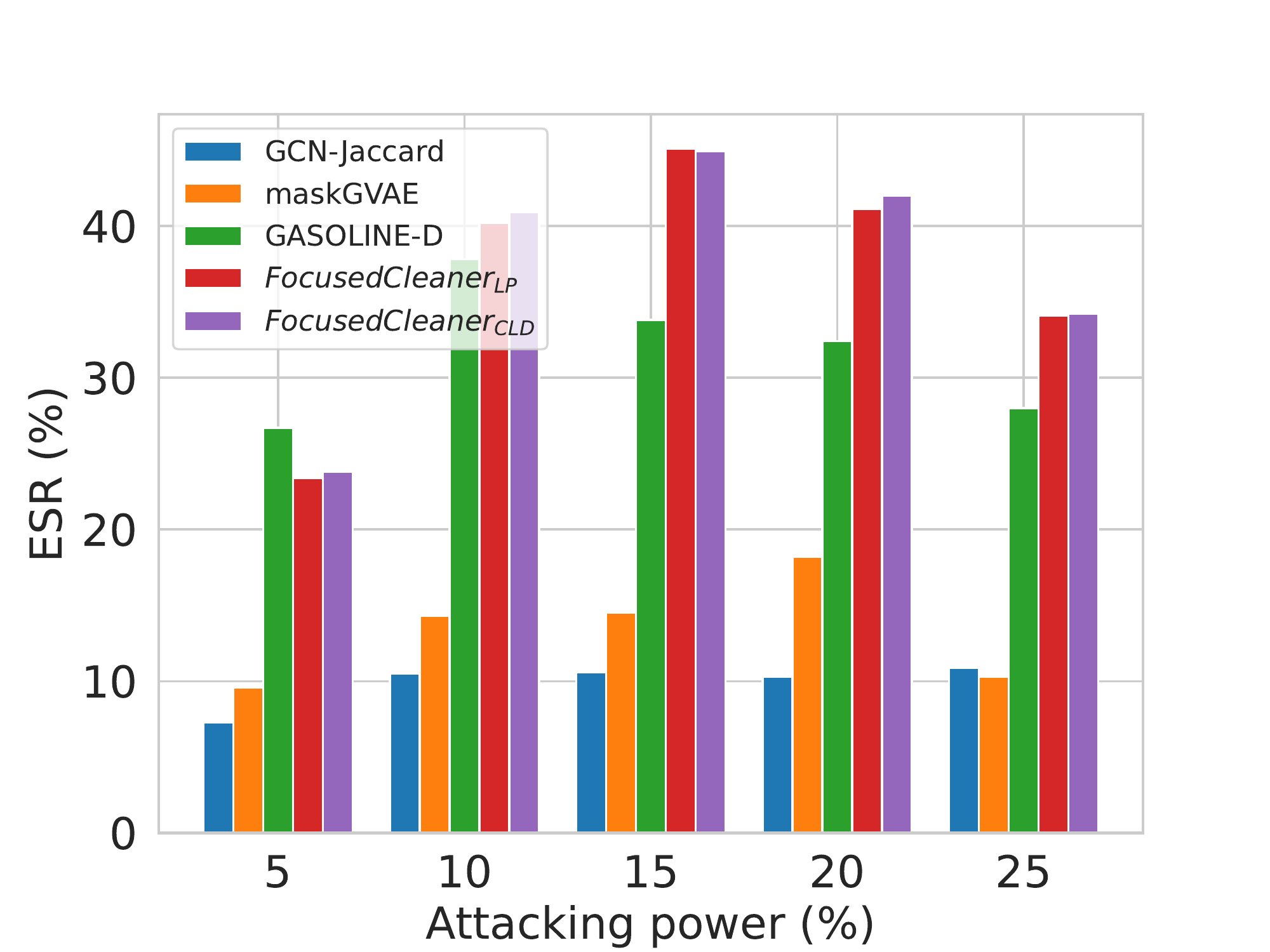}
		\caption{Citeseer for MinMax}
	\end{subfigure}
    \hfill
	\begin{subfigure}[b]{0.33\textwidth}
		\centering
		\includegraphics[width=\textwidth,height=3.5cm]{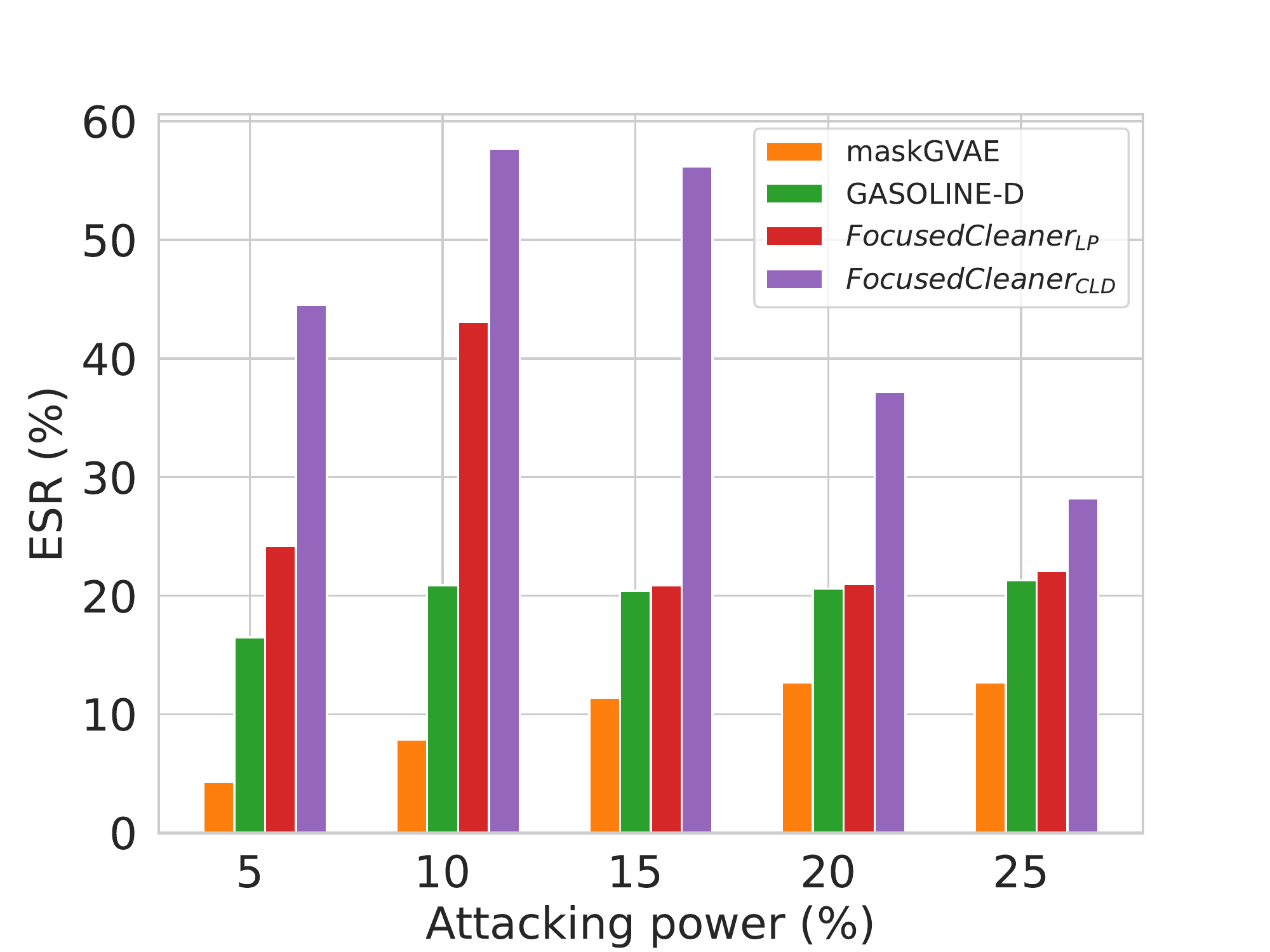}
		\caption{Polblogs for MinMax}
	\end{subfigure}
	\caption{Sanitation results against two attacks based on $\mathsf{ESR}$.}
	\label{fig-rec-attack}
\end{figure*}

\begin{figure*}[htp]
	\centering
        \begin{subfigure}[b]{0.33\textwidth}
		\centering
		\includegraphics[width=\textwidth,height=3.5cm]{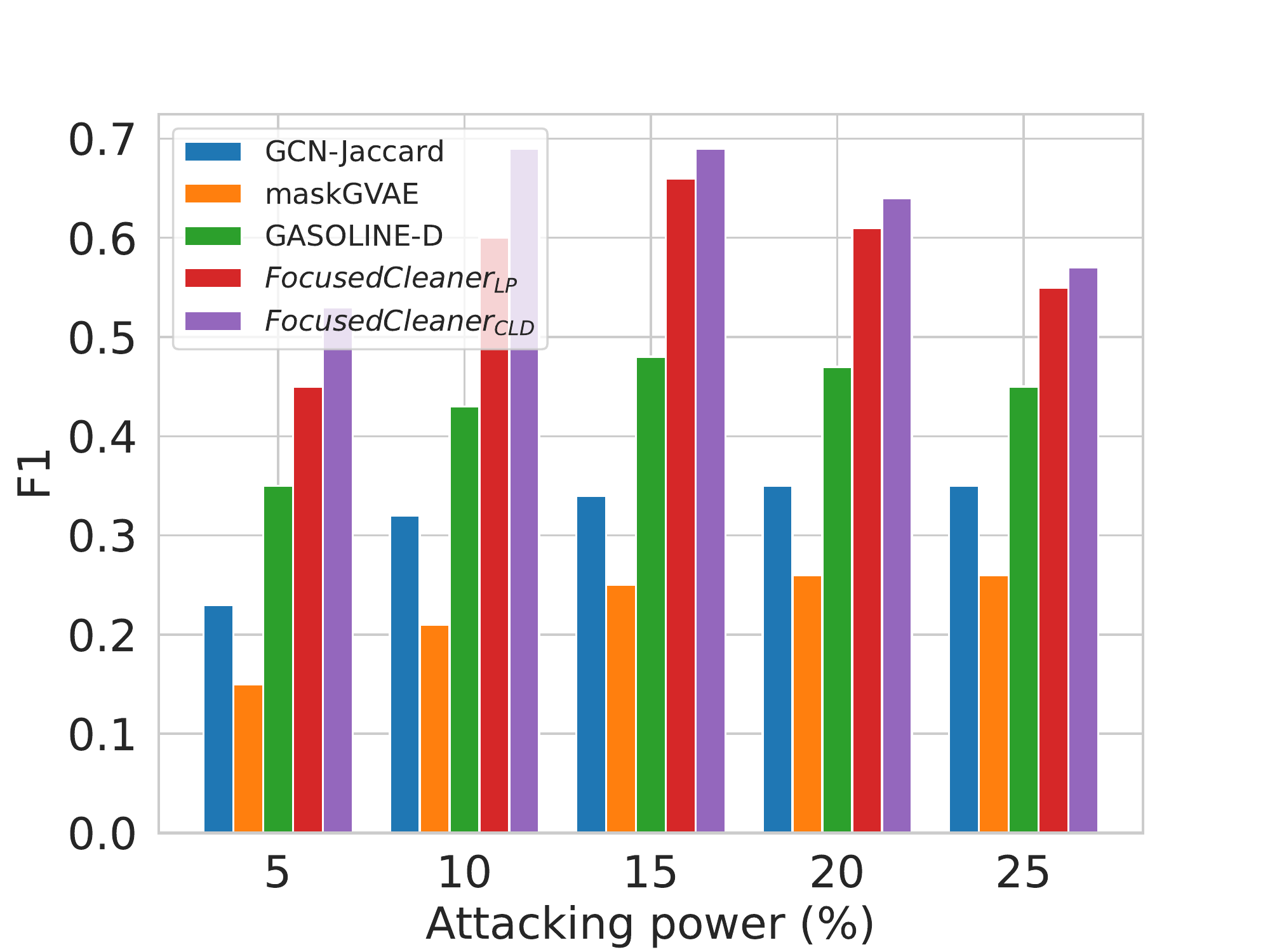}
		\caption{Cora for METTACK}
	\end{subfigure}
        \hfill
	\begin{subfigure}[b]{0.33\textwidth}
		\centering
		\includegraphics[width=\textwidth,height=3.5cm]{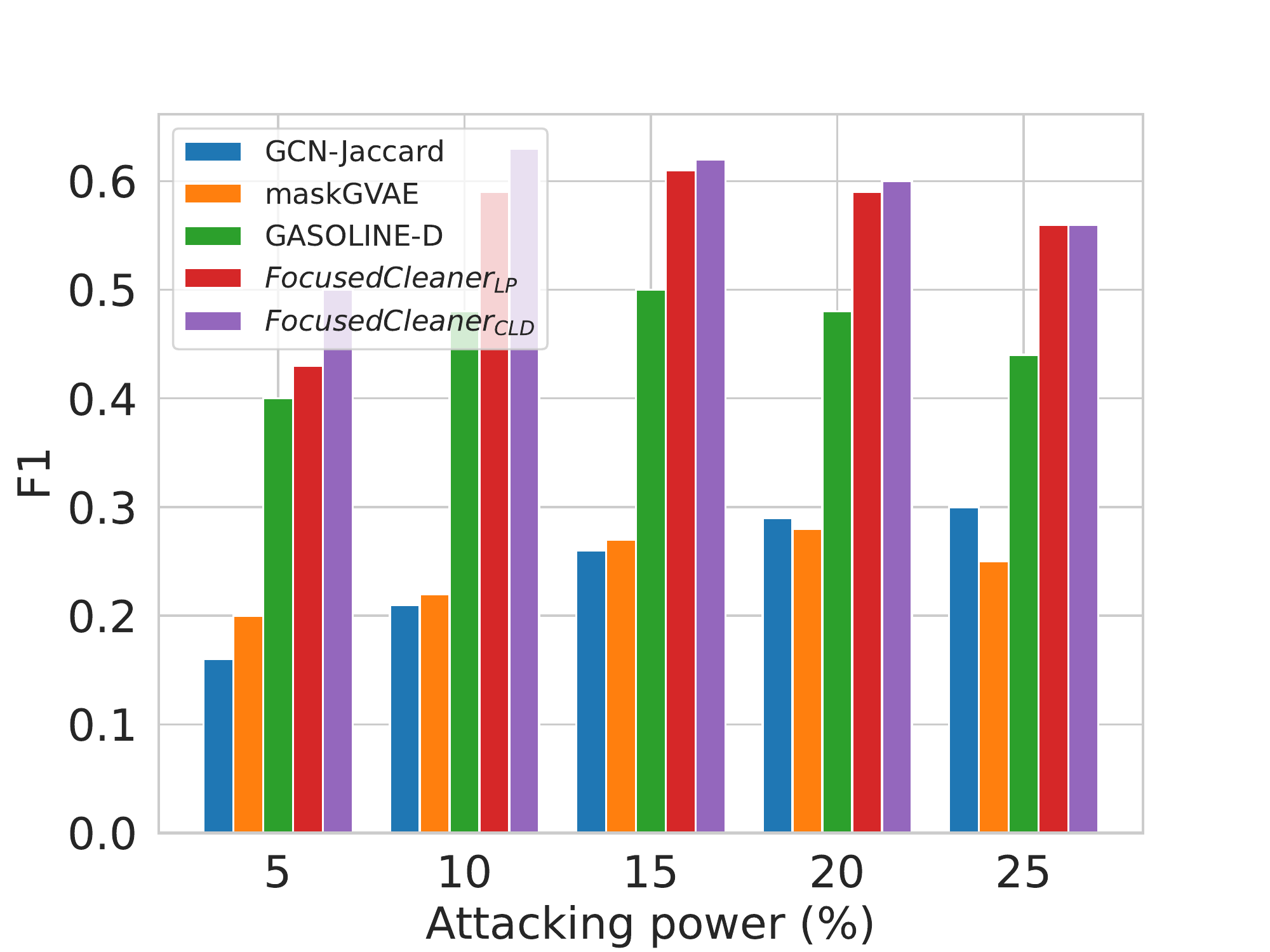}
		\caption{Citeseer for METTACK}
	\end{subfigure}
        \hfill
	\begin{subfigure}[b]{0.33\textwidth}
		\centering
		\includegraphics[width=\textwidth,height=3.5cm]{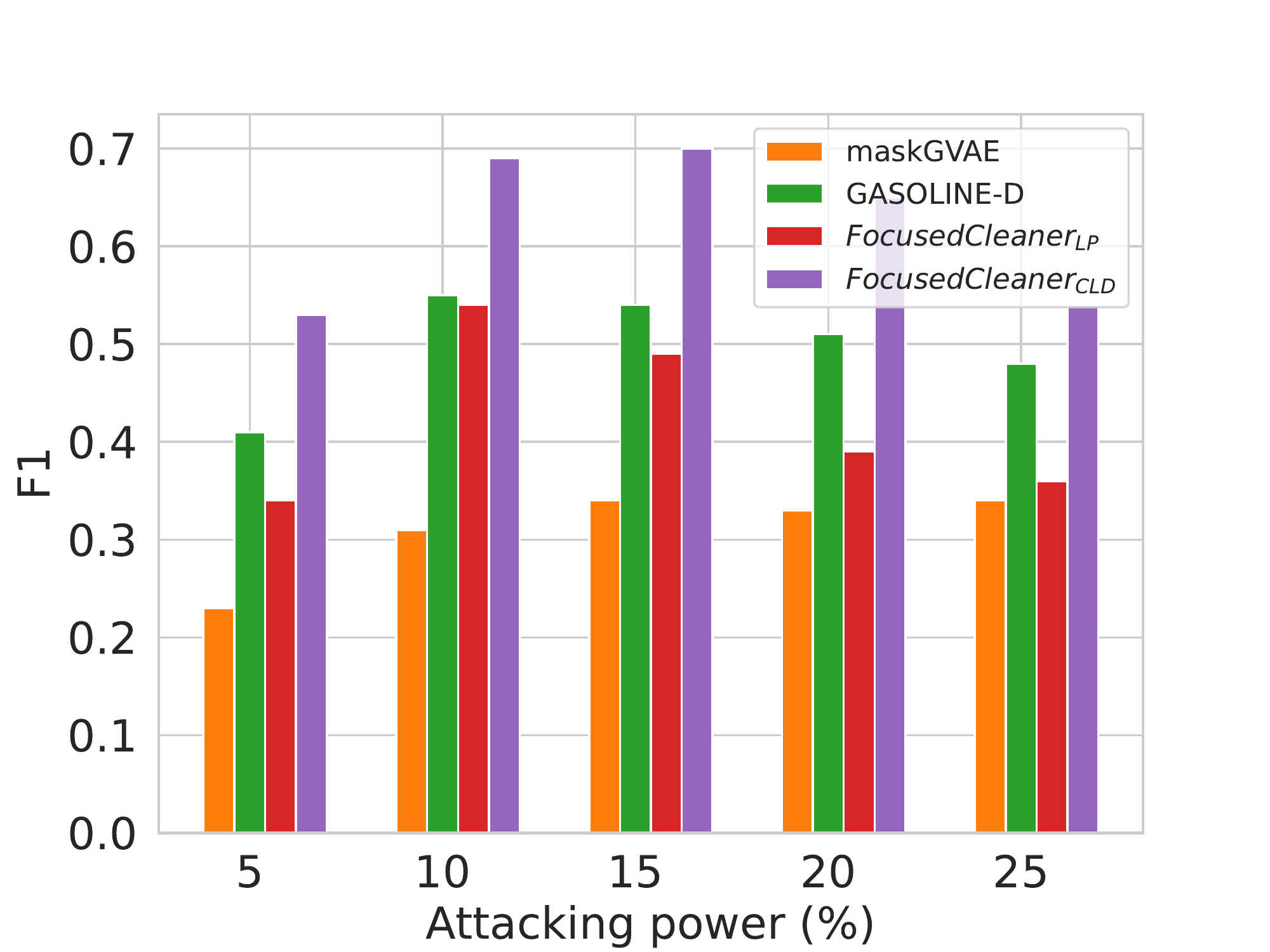}
		\caption{Polblogs for METTACK}
	\end{subfigure}
	\hfill
	\begin{subfigure}[b]{0.33\textwidth}
		\centering
		\includegraphics[width=\textwidth,height=3.5cm]{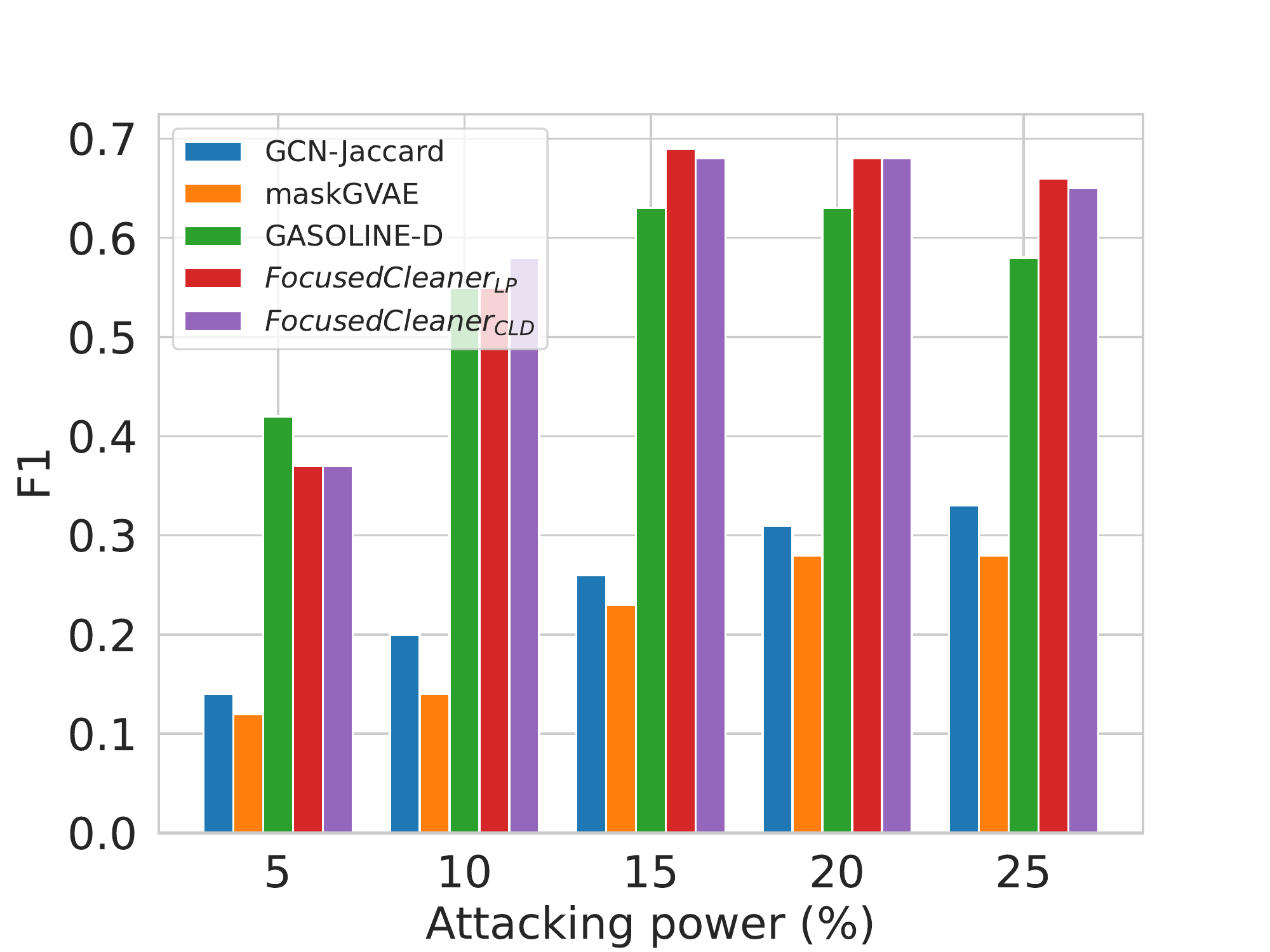}
		\caption{Cora for MinMax}
	\end{subfigure}
	\hfill
	\begin{subfigure}[b]{0.33\textwidth}
		\centering
		\includegraphics[width=\textwidth,height=3.5cm]{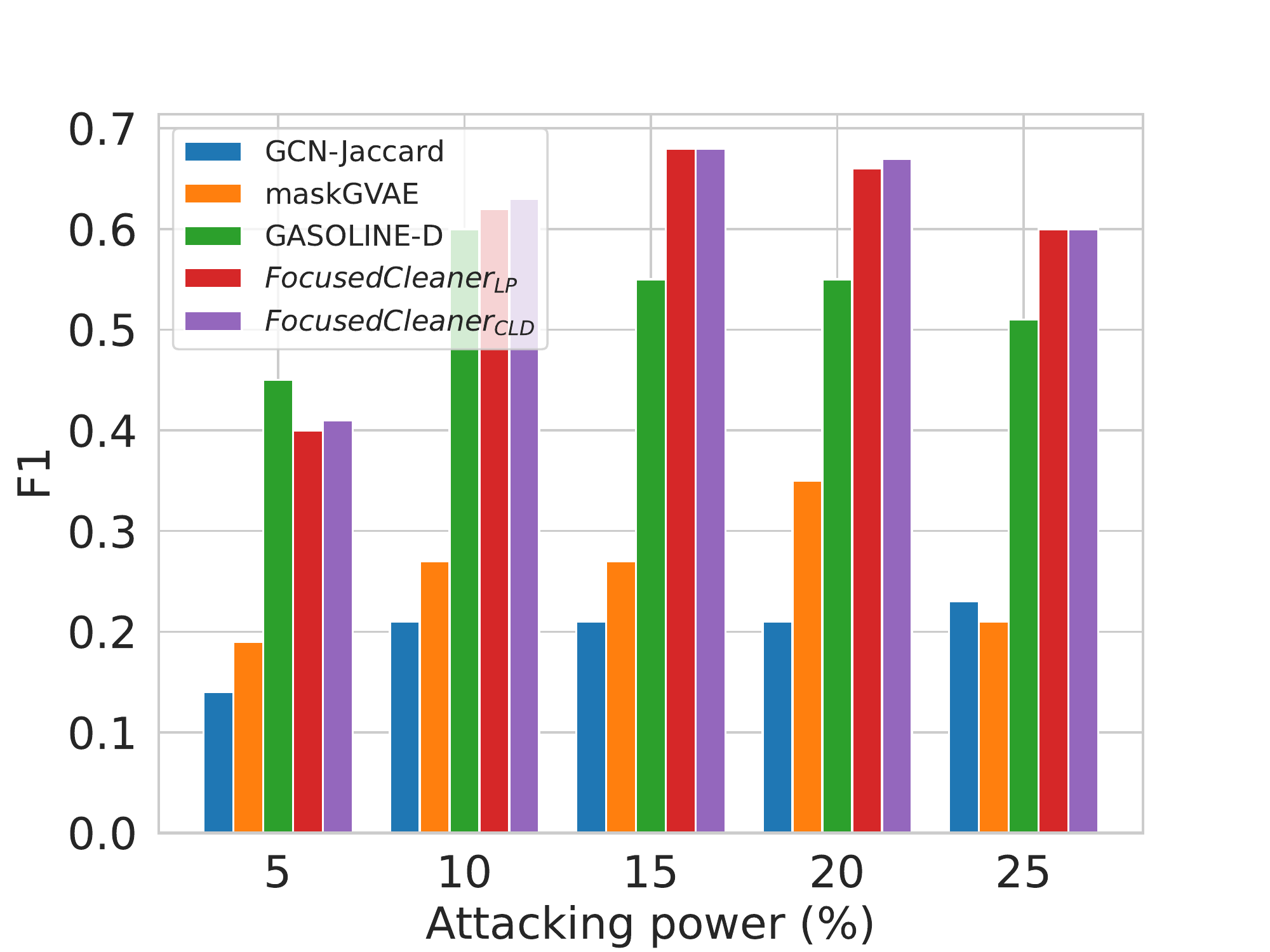}
		\caption{Citeseer for MinMax}
	\end{subfigure}
	\hfill
	\begin{subfigure}[b]{0.33\textwidth}
		\centering
		\includegraphics[width=\textwidth,height=3.5cm]{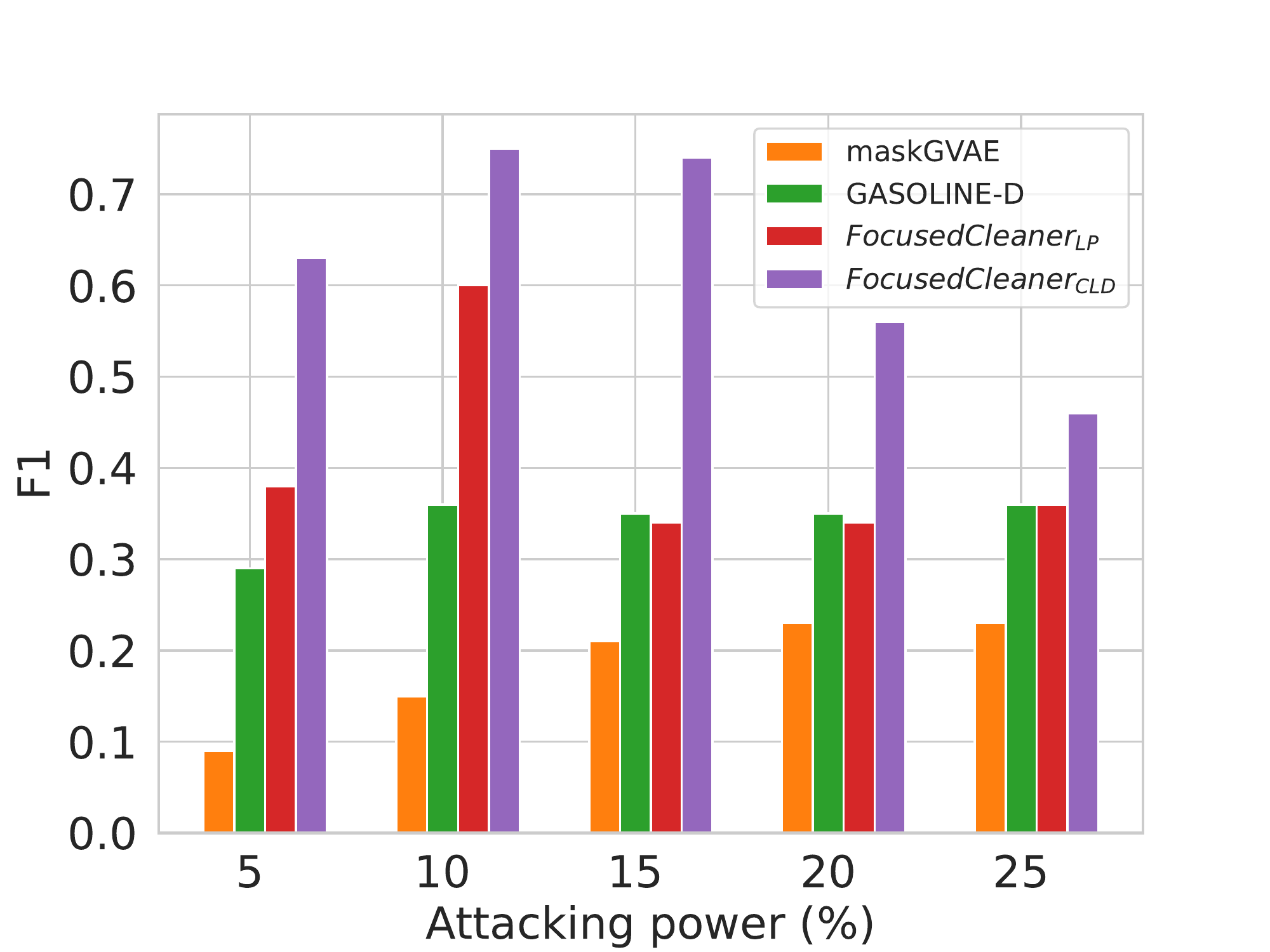}
		\caption{Polblogs for MinMax}
	\end{subfigure}
	\caption{Sanitation results against two attacks based on $F_1$ score.}
	\label{fig-f1-attack}
\end{figure*}

\subsubsection{Sanitation Results}
Since we have two different victim node detectors \textbf{ClassDiv} and \textbf{LinkPred}, we term the corresponding sanitizer as \textsf{FocusedCleaner}$_{CLD}$ and \textsf{FocusedCleaner}$_{LP}$, respectively.  We use the attacks to poison a graph with varying \textit{attack power}, which is the fraction of malicious edges over all edges in the clean graph. We note that the attacks are allowed to insert and remove edges; however, we observe that more than $99\%$ of operations are inserting.

Fig.~\ref{fig-rec-attack} and \ref{fig-f1-attack} presents the $\mathsf{ESR}$ and $F_1$ score of different sanitizers against METTACK and MinMax under various attack powers. Note, the results of GCN-Jaccard are missing on Polblogs because this method relies on node features. It is observed that our sanitizers outperform others by a large margin in almost all cases, which demonstrates the effectiveness of the detection module in providing a focus for sanitation. Overall, \textsf{FocusedCleaner}$_{CLD}$ is slightly better than \textsf{FocusedCleaner}$_{LP}$. We also observe that on Polblogs, our sanitizers are not as good as GASOLINE-D in some cases. The possible reason is that nodes in Polblogs have no features, which might impair the victim node detection's performance. 
%In addition, we also use the standard $F_1$ score as the metric. In our context,  $F_1 =2|\mathcal{S}_{atk} \cap \mathcal{S}_{san}|/(|\mathcal{S}_{atk}| + |\mathcal{S}_{san}|)$. 
%We leave the sanitation results based on F1 score in the supplement. 
It is observed that $F_1$ scores show a similar trend as $\mathsf{ESR}$ and our methods almost always outperform other baselines. 

\begin{table*}[h]
\centering
\caption{Mean node classification accuracy (\%) under different attacking powers for METTACK.}
\label{tab-defense-METTACK}
\resizebox{2.1\columnwidth}{!}{%
\begin{tabular}{l|rrrrrrr|rrrrrrr}
\toprule[1.pt]
\multicolumn{8}{c|}{Preprocessing-based methods}&\multicolumn{7}{c}{Robust models}\\
\hline
Dataset& Ptb rate & GCN & GCN-Jaccard & maskGVAE & GASOLINE-D & \textsf{FocusedCleaner}$_{LP}$ & \textsf{FocusedCleaner}$_{CLD}$ & RGCN& ProGNN & MedianGNN & SimPGCN & GNNGUARD & ElasticGNN & AirGNN  \\
\multirow{5}*{Cora}&$5\%$&$71.1$&$75.2$&$74.4$&$79.4$&$81.0$&$\textbf{83.6}$&$71.6$&$76.3$&$77.0$&$70.6$&$71.6$&$75.4$&$70.8$\\
                  &$10\%$&$61.4$&$68.2$&$63.0$&$74.1$&$79.3$&$\textbf{81.1}$&$63.5$&$72.2$&$71.5$&$61.8$&$61.8$&$67.1$&$61.7$\\
                  &$15\%$&$52.8$&$60.5$&$57.2$&$66.2$&$\textbf{76.5}$&$76$&$56.3$&$65.6$&$64.5$&$55.1$&$54.3$&$59.5$&$52.5$\\
                  &$20\%$&$42.8$&$54.3$&$50.2$&$58.2$&$71.1$&$\textbf{71.9}$&$52.9$&$61.9$&$60.3$&$50.8$&$47.6$&$53.9$&$43.9$\\
                  &$25\%$&$35.0$&$47.5$&$43.7$&$51.5$&$\textbf{66.0}$&$\textbf{66.0}$&$48.3$&$58.8$&$57.1$&$46.0$&$40.4$&$47.3$&$36.4$\\
\hline
\multirow{5}*{Citeseer}&$5\%$&$62.8$&$63.9$&$63.8$&$69.8$&$71.6$&$\textbf{74.9}$&$61.9$&$70.6$&$64.0$&$63.8$&$63.1$&$65.1$&$62.7$\\
                      &$10\%$&$55.7$&$56.9$&$57.4$&$62.6$&$68.7$&$\textbf{71.7}$&$55.2$&$65.0$&$58.6$&$56.4$&$55.2$&$58.1$&$55.8$\\
                      &$15\%$&$49.8$&$50.5$&$52.1$&$58.5$&$64.9$&$\textbf{65.5}$&$50.1$&$60.4$&$55.1$&$50.3$&$48.9$&$52.4$&$49.5$\\
                      &$20\%$&$40.7$&$48.7$&$43.0$&$53.2$&$\textbf{60.8}$&$60.3$&$44.8$&$53.5$&$50.7$&$47.4$&$42.9$&$47.6$&$43.2$\\
                      &$25\%$&$36.7$&$39.4$&$40.6$&$48.7$&$\textbf{58.0}$&$56.1$&$40.1$&$48.2$&$47.2$&$44.7$&$37.1$&$42.3$&$37.4$\\
\hline
\multirow{5}*{Polblogs}&$5\%$&$79.1$&$/$   &$82.9$&$80.8$&$84.8$&$\textbf{88.5}$&$80.0$&$87.8$&$85.6$&$52.4$&$77.2$&$83.3$&$58.9$\\
                      &$10\%$&$71.9$&$/$   &$76.1$&$81.7$&$83.4$&$\textbf{88.0}$&$72.0$&$79.0$&$78.7$&$52.2$&$70.5$&$77.2$&$57.1$\\
                      &$15\%$&$67.5$&$/$   &$71.1$&$76.0$&$78.8$&$\textbf{86.0}$&$67.2$&$72.4$&$73.0$&$51.5$&$66.2$&$71.2$&$55.4$\\
                      &$20\%$&$66.7$&$/$   &$68.4$&$73.9$&$74.4$&$\textbf{78.9}$&$66.1$&$70.1$&$70.4$&$51.2$&$65.5$&$69.0$&$55.2$\\
                      &$25\%$&$66.2$&$/$   &$66.5$&$71.5$&$\textbf{72.3}$&$72.2$&$65.6$&$67.6$&$68.1$&$50.8$&$65.3$&$68.5$&$55.2$\\
\bottomrule[1.pt]             
\end{tabular}
}
\end{table*}

\begin{table*}[h]
\centering
\caption{Mean node classification accuracy (\%) under different attacking powers for MinMax.}
\label{tab-defense-MinMax}
\resizebox{2.1\columnwidth}{!}{%
\begin{tabular}{l|rrrrrrr|rrrrrrr}
\toprule[1.pt]
\multicolumn{8}{c|}{Preprocessing-based methods}&\multicolumn{7}{c}{Robust models}\\
\hline
Dataset& Ptb rate & GCN & GCN-Jaccard & maskGVAE & GASOLINE-D & \textsf{FocusedCleaner}$_{LP}$ & \textsf{FocusedCleaner}$_{CLD}$ & RGCN& ProGNN & MedianGNN & SimPGCN & GNNGUARD & ElasticGNN & AirGNN  \\
\multirow{5}*{Cora}&$5\%$&$76.9$&$76.4$&$77.5$&$83.3$&$\textbf{83.6}$&$83.0$&$78.9$&$81.3$&$79.4$&$81.3$&$76.1$&$80.7$&$75.3$\\
                  &$10\%$&$75.4$&$74.9$&$75.2$&$80.6$&$82.3$&$\textbf{82.9}$&$77.0$&$77.0$&$75.0$&$80.1$&$75.1$&$81.0$&$70.7$\\
                  &$15\%$&$70.6$&$71.3$&$70.9$&$76.6$&$\textbf{81.8}$&$81.2$&$72.1$&$71.1$&$69.1$&$77.7$&$69.5$&$77.5$&$64.5$\\
                  &$20\%$&$63.9$&$67.3$&$65.6$&$75.0$&$79.5$&$\textbf{79.6}$&$64.6$&$58.2$&$57.5$&$74.9$&$61.9$&$75.1$&$60.4$\\
                  &$25\%$&$58.6$&$63.5$&$58.4$&$68.2$&$\textbf{77.4}$&$75.7$&$56.4$&$47.0$&$48.0$&$73.4$&$57.5$&$72.8$&$49.4$\\
\hline
\multirow{5}*{Citeseer}&$5\%$&$69.4$&$69.0$&$69.5$&$73.1$&$73.9$&$\textbf{74.1}$&$66.7$&$70.1$&$71.1$&$73.0$&$68.7$&$72.4$&$64.4$\\
                      &$10\%$&$66.7$&$67.6$&$68.0$&$70.7$&$\textbf{73.7}$&$73.5$&$65.3$&$68.5$&$68.4$&$73.3$&$63.8$&$72.5$&$56.9$\\
                      &$15\%$&$65.7$&$64.8$&$66.1$&$70.4$&$\textbf{73.2}$&$73.0$&$64.1$&$68.0$&$66.3$&$71.9$&$64.6$&$70.8$&$54.5$\\
                      &$20\%$&$64.1$&$65.0$&$65.8$&$68.7$&$\textbf{73.2}$&$73.1$&$60.4$&$63.8$&$65.0$&$71.8$&$63.2$&$71.2$&$51.3$\\
                      &$25\%$&$57.9$&$59.0$&$58.9$&$64.1$&$69.3$&$69.3$&$54.1$&$55.0$&$55.4$&$\textbf{71.1}$&$57.6$&$68.5$&$47.4$\\
\hline
\multirow{5}*{Polblogs}&$5\%$&$89.0$&$/$&$90.0$&$71.4$&$90.0$&$90.9$&$90.5$&$91.8$&$85.6$&$59.6$&$87.2$&$\textbf{92.2}$&$78.2$\\
                      &$10\%$&$77.2$&$/$&$70.9$&$60.3$&$87.9$&$\textbf{90.9}$&$69.2$&$85.8$&$56.8$&$52.8$&$76.5$&$90.8$&$62.6$\\
                      &$15\%$&$62.6$&$/$&$73.6$&$54.5$&$67.0$&$\textbf{90.3}$&$52.8$&$81.4$&$19.1$&$52.6$&$62.2$&$77.3$&$56.7$\\
                      &$20\%$&$53.8$&$/$&$\textbf{66.8}$&$51.3$&$55.5$&$63.4$&$53.8$&$63.0$&$19.3$&$51.8$&$57.0$&$65.2$&$51.4$\\
                      &$25\%$&$52.8$&$/$&$\textbf{69.2}$&$49.9$&$56.0$&$67.2$&$52.3$&$68.8$&$39.6$&$51.4$&$67.8$&$63.9$&$51.4$\\
\bottomrule[1.pt]             
\end{tabular}
}
\end{table*}

\subsubsection{Discussion on Affordable Sanitation Budget $\mathcal{R}_{ASB}$}
\begin{figure}[h]
	\centering
	\begin{subfigure}[b]{0.236\textwidth}
		\centering
		\includegraphics[width=\textwidth,height=3.1cm]{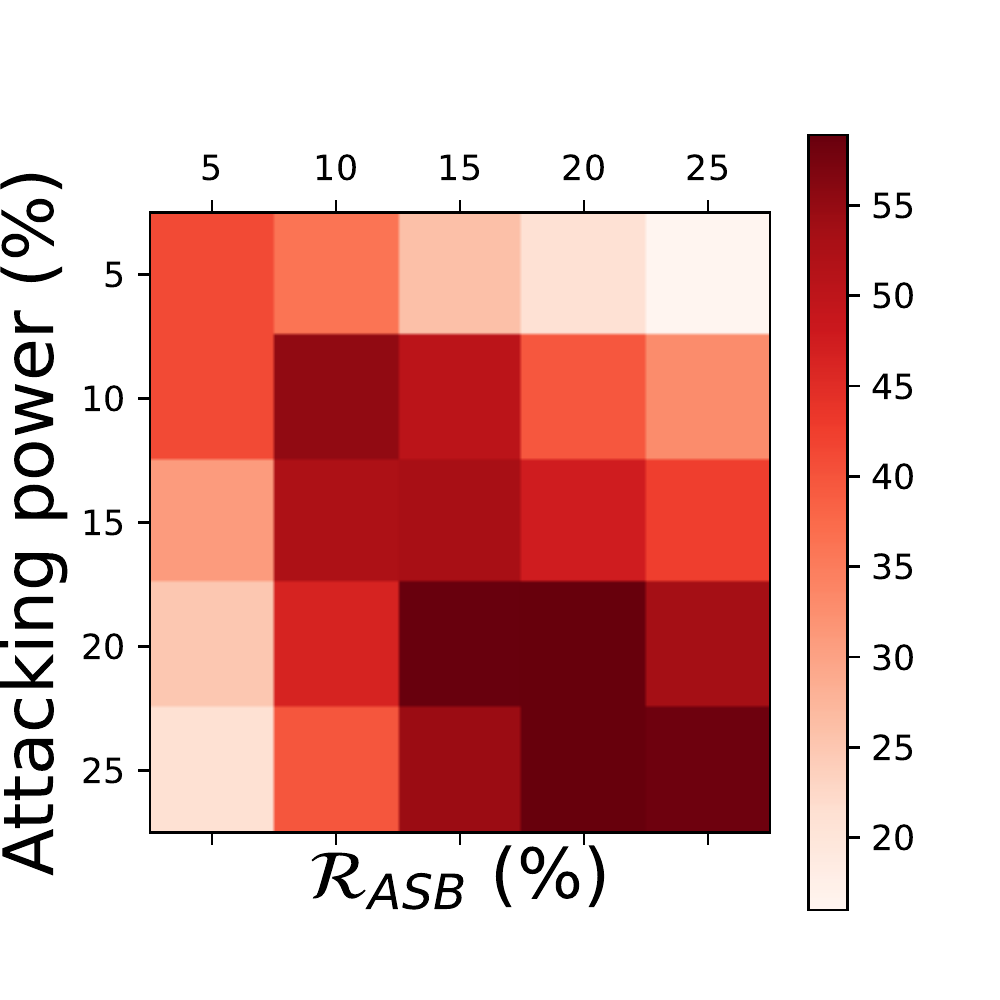}
		\caption{Cora}
	\end{subfigure}
	\hfill
	\begin{subfigure}[b]{0.236\textwidth}
		\centering
		\includegraphics[width=\textwidth,height=3.1cm]{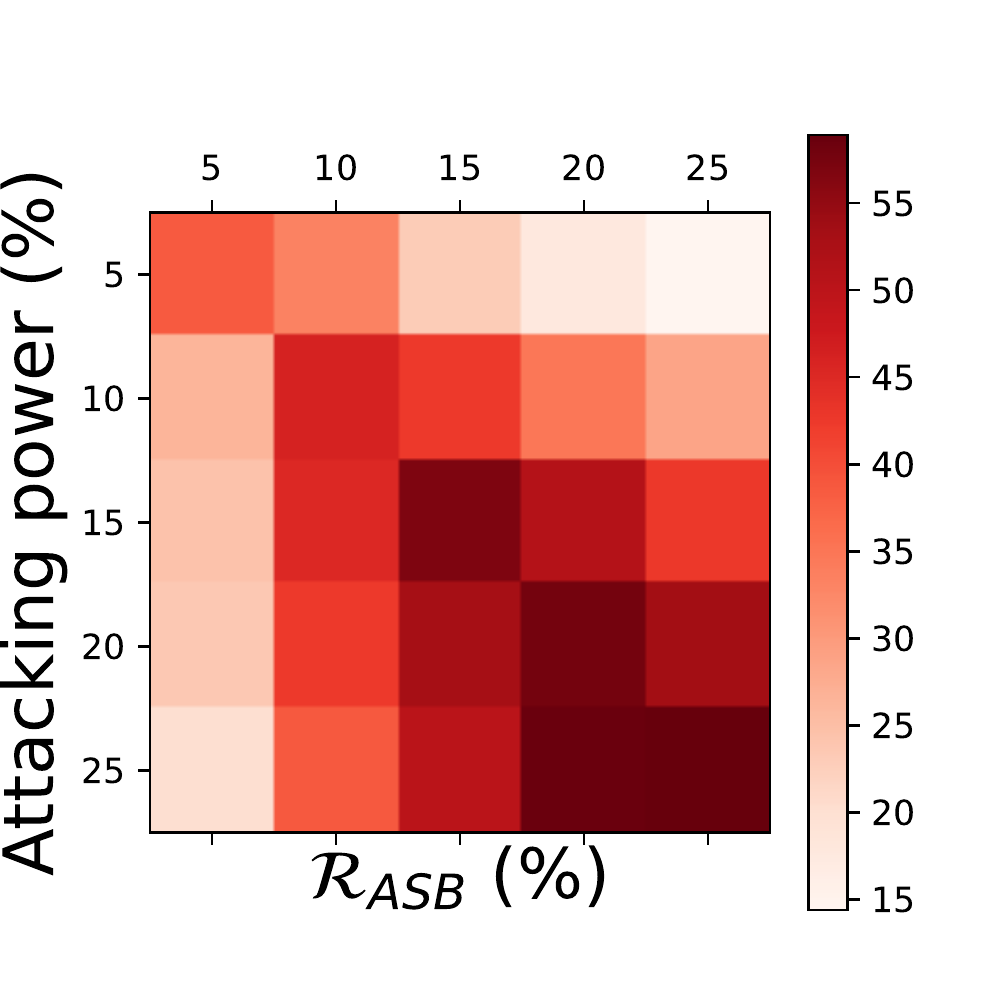}
		\caption{Citeseer}
	\end{subfigure}
	\caption{$\mathsf{ESR}$ with varying $\mathcal{R}_{ASB}$ and attacking powers.}
	\label{fig-atk-vs-san}
\end{figure}

\begin{figure}[h]
	\centering
	\begin{subfigure}[b]{0.234\textwidth}
		\centering
		\includegraphics[width=\textwidth,height=2.8cm]{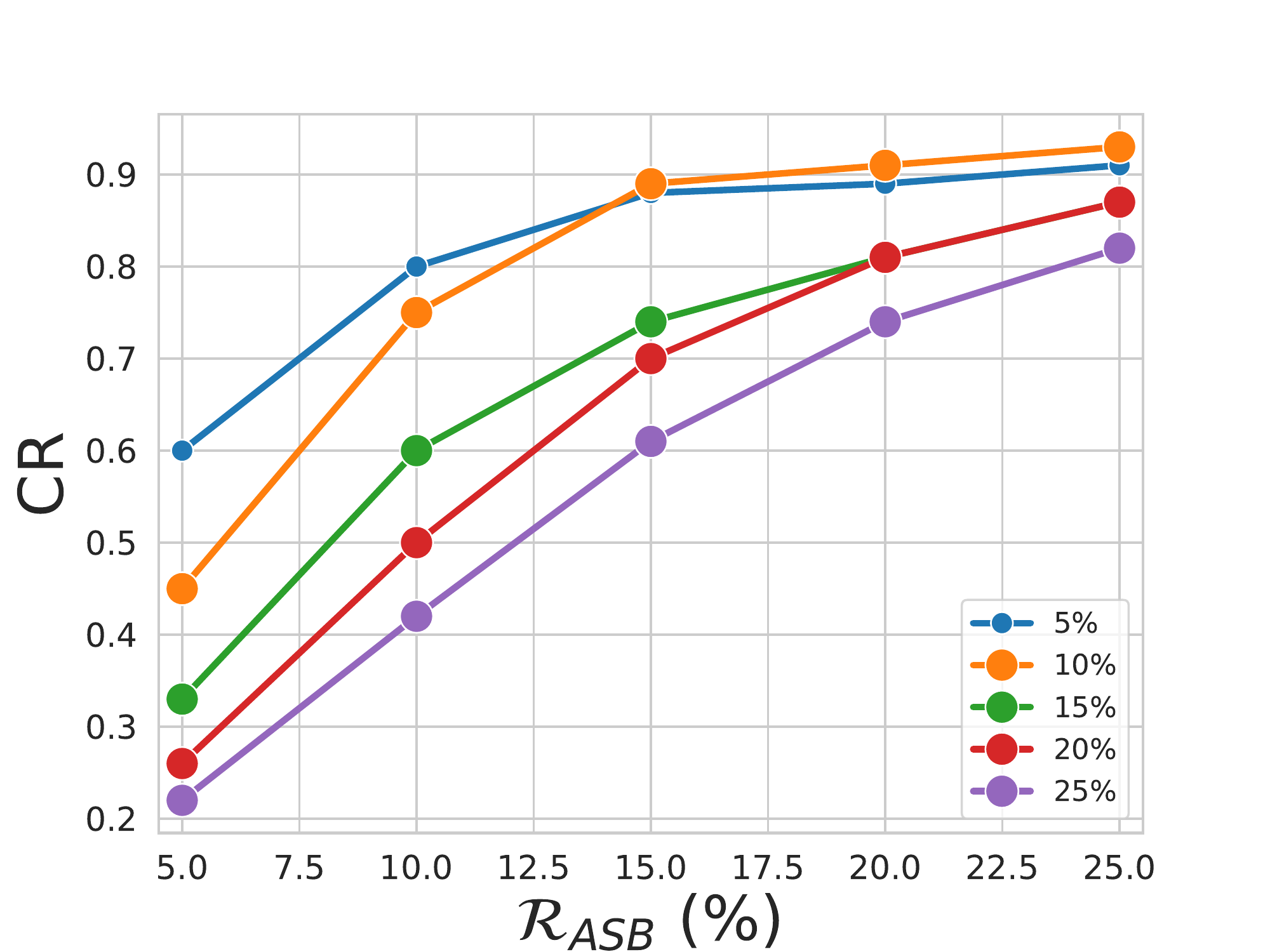}
		\caption{Cora}
	\end{subfigure}
	\hfill
	\begin{subfigure}[b]{0.234\textwidth}
		\centering
		\includegraphics[width=\textwidth,height=2.8cm]{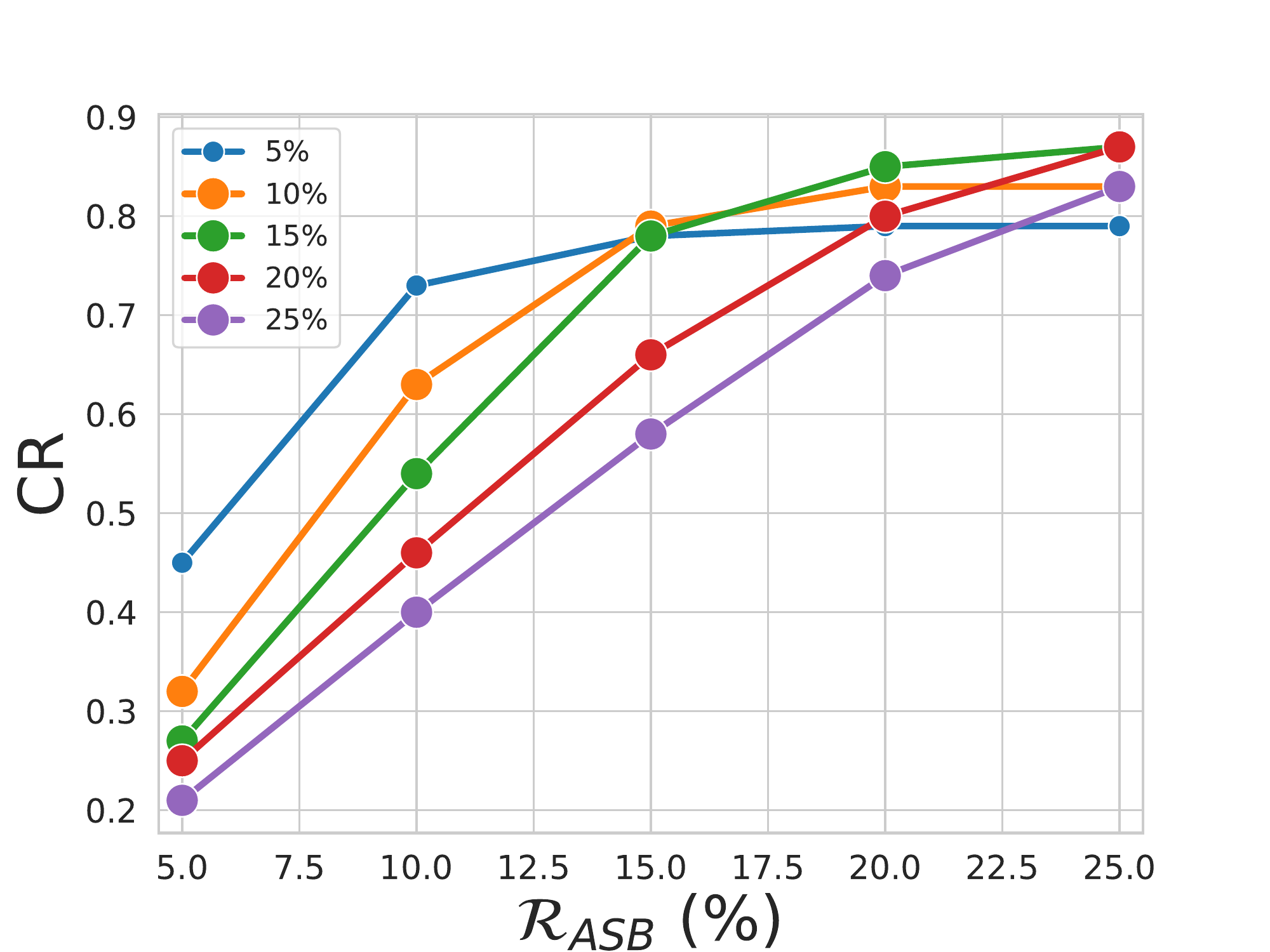}
		\caption{Citeseer}
	\end{subfigure}
	\caption{$\mathsf{CR}$ with different $\mathcal{R}_{ASB}$ and attacking powers.}
	\label{fig-atk-vs-cr}
\end{figure}

The selection of $\mathcal{R}_{ASB}$ is a practical issue. It depends on the defender's ability to process the sanitation results as well as his prior knowledge of the attack power, i.e., the number of malicious edges expected in the poisoned graph.
Thus, it is beneficial to see the effects of different levels of $\mathcal{R}_{ASB}$ on sanitation performance. To this end, we vary the values of $\mathcal{R}_{ASB}$ and check the performance of the sanitizer against different attack powers. We use  \textsf{FocusedCleaner}$_{CLD}$ against METTACK as the example. The sanitation results ($\mathsf{ESR}$) are shown in Fig.~\ref{fig-atk-vs-san}. In general, when $\mathcal{R}_{ASB}$ matches the attack power (the diagonal entries), we can achieve the best recovery ratios. In addition, Fig.~\ref{fig-atk-vs-cr} shows the values of  $\mathsf{CR}$ as we increase the budget $\mathcal{R}_{ASB}$. As expected, as the defender invests more (i.e., larger $\mathcal{R}_{ASB}$), more malicious edges (over $80\%$) will be identified for all attack powers. Moreover, for a fixed attack power (i.e., a particular line in Fig.~\ref{fig-atk-vs-cr}), the increase of $\mathsf{CR}$ will slow down as we increase $\mathcal{R}_{ASB}$, demonstrating a typical \textit{diminishing return} phenomenon -- it is more evident for lower attack powers (i.e., fewer injected malicious edges).

%It is observed that the coverage ratio becomes larger when we increase our budget $\mathcal{R}_{ASB}$. This phenomenon demonstrates that if the defender has unlimited resources and utilizes the sanitizer to identify maliciously adversarial edges, it can cover up to around $90\%$ of the adversarial edges injected by the attacker.

%At the same time, we observe that when the sanitation ratio is set to be around $10\%$ to $15\%$, the sanitation results are satisfactory in general under all attack powers.

\subsection{Sensitivity Analysis}
\begin{figure}[htp]
	\centering
	\begin{subfigure}[b]{0.236\textwidth}
		\centering
		\includegraphics[width=\textwidth,height=2.8cm]{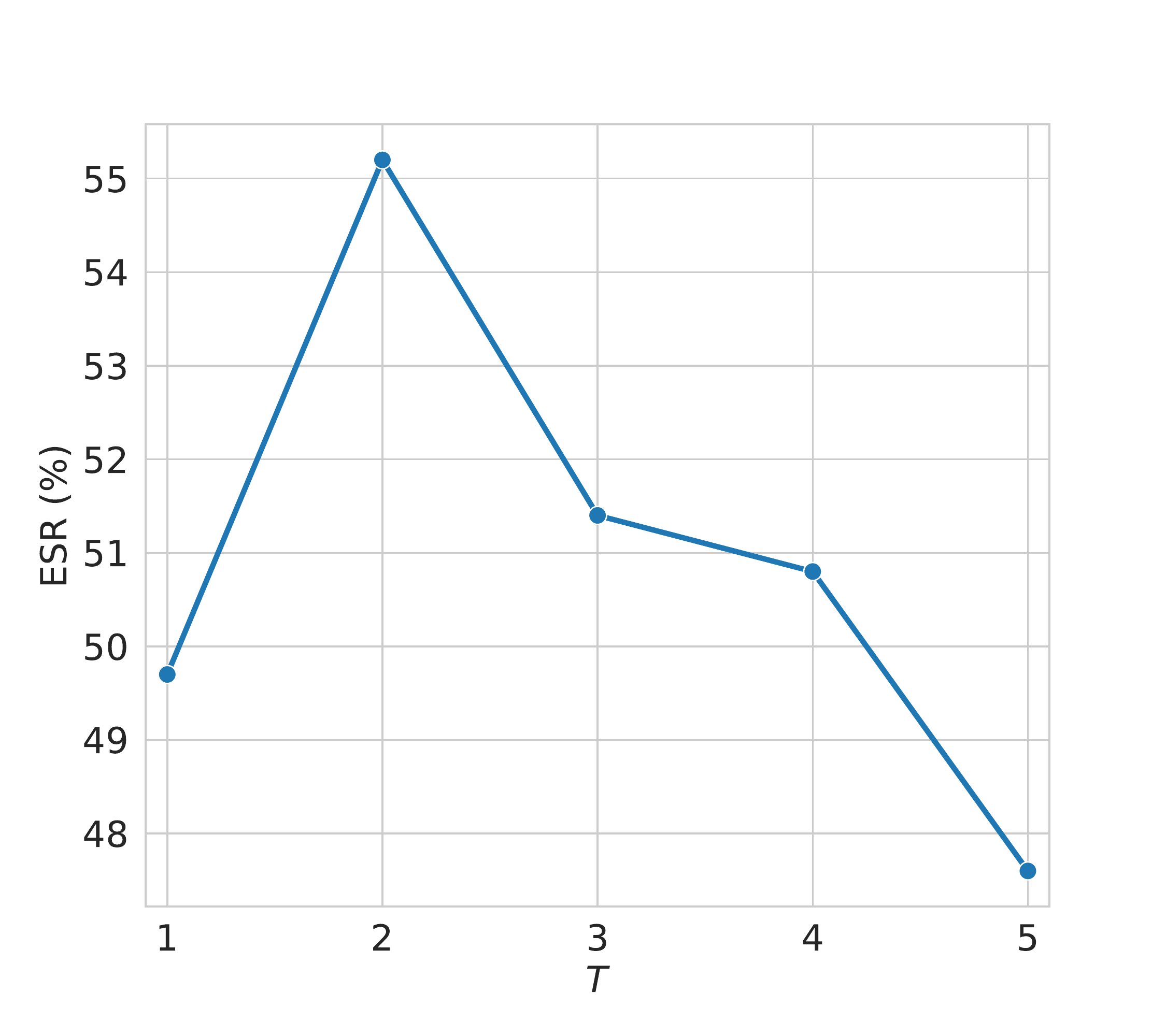}
		\caption{$T$}
		\label{fig-sensitivity-T}
	\end{subfigure}
	\hfill
	\begin{subfigure}[b]{0.236\textwidth}
		\centering
		\includegraphics[width=\textwidth,height=2.8cm]{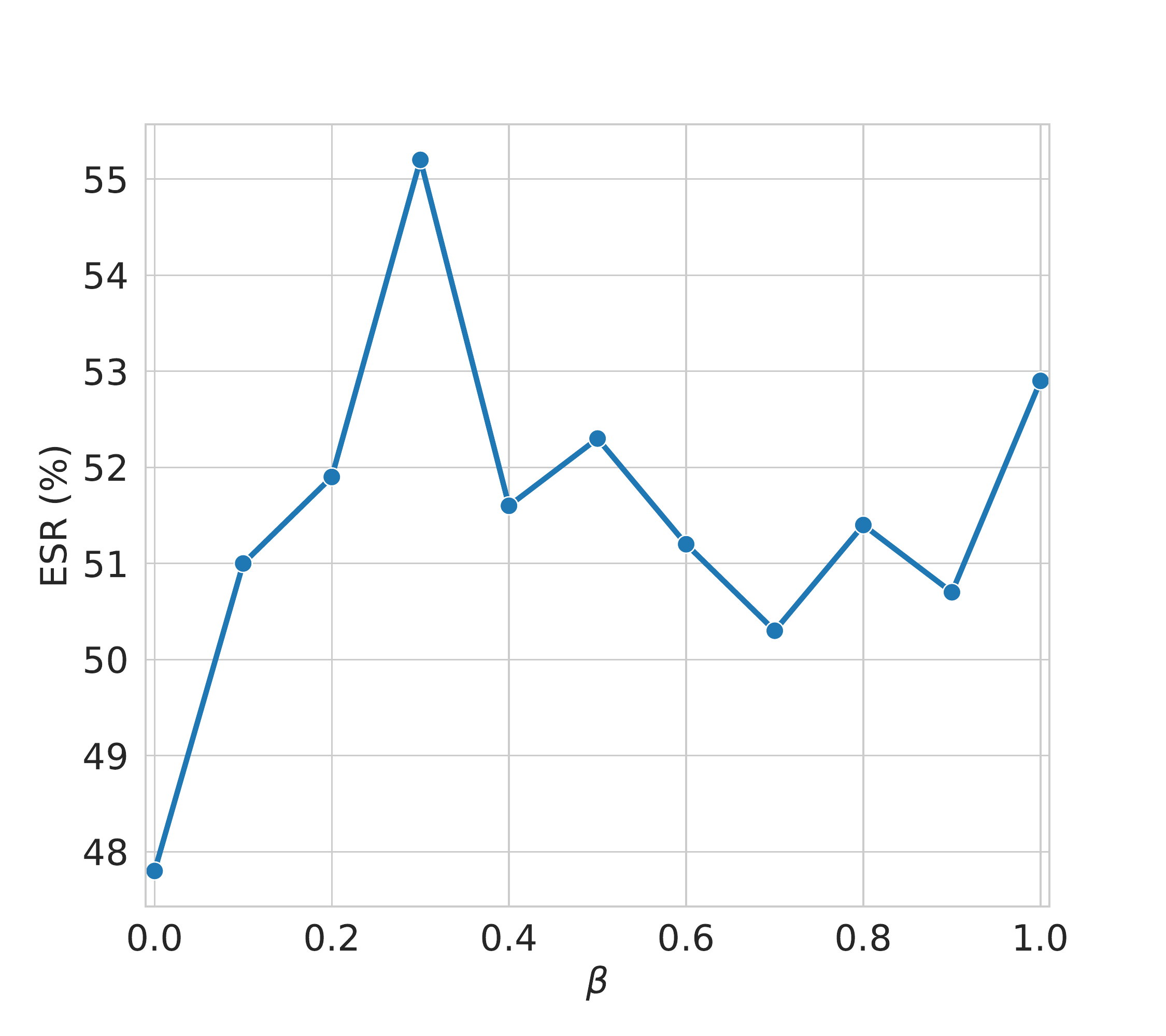}
		\caption{$\beta$}
		\label{fig-sensitivity-beta}
	\end{subfigure}
	\hfill
	\begin{subfigure}[b]{0.236\textwidth}
		\centering
		\includegraphics[width=\textwidth,height=2.8cm]{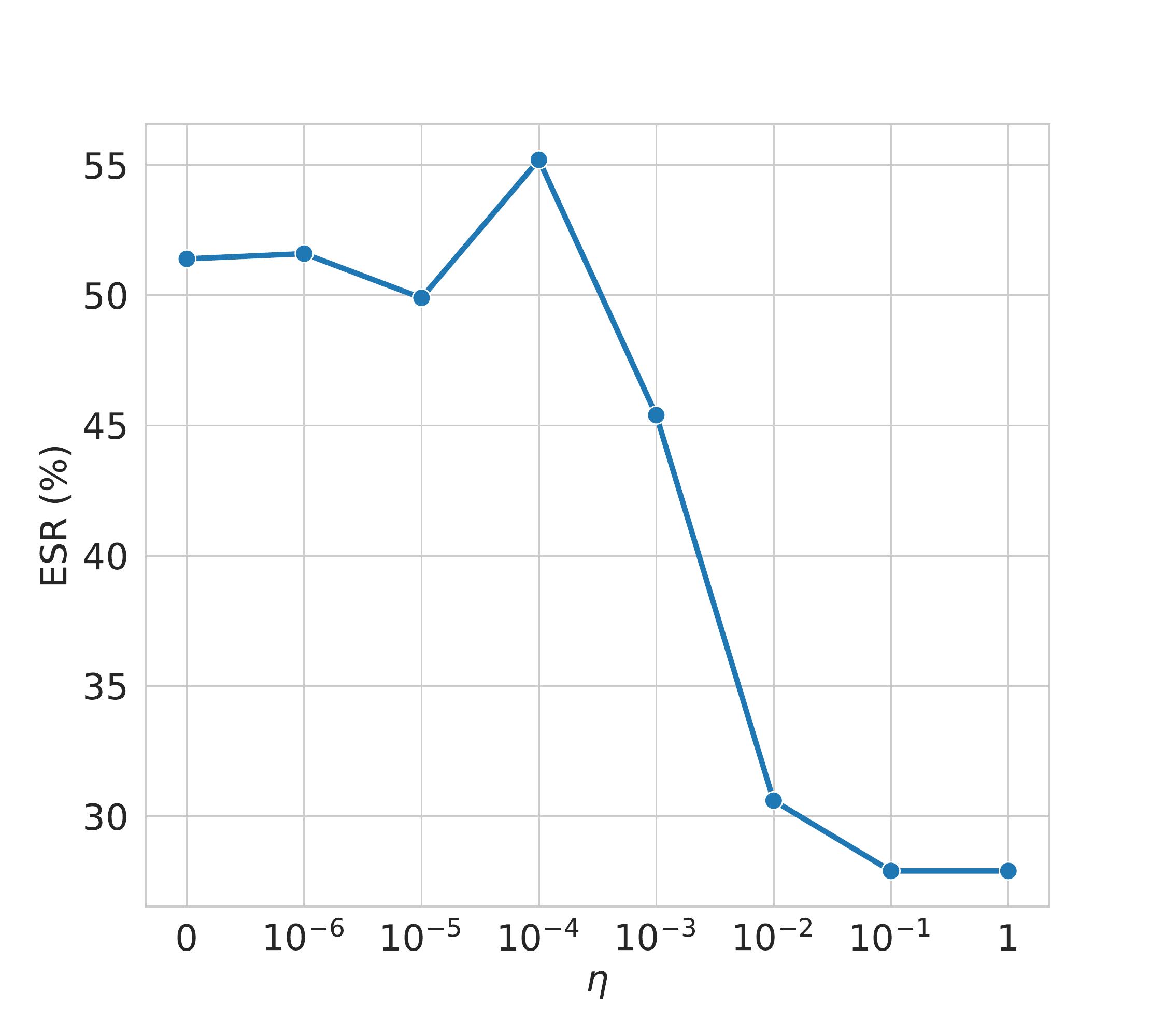}
		\caption{$\eta$}
		\label{fig-sensitivity-eta}
	\end{subfigure}
	\hfill
	\begin{subfigure}[b]{0.236\textwidth}
		\centering
		\includegraphics[width=\textwidth,height=2.8cm]{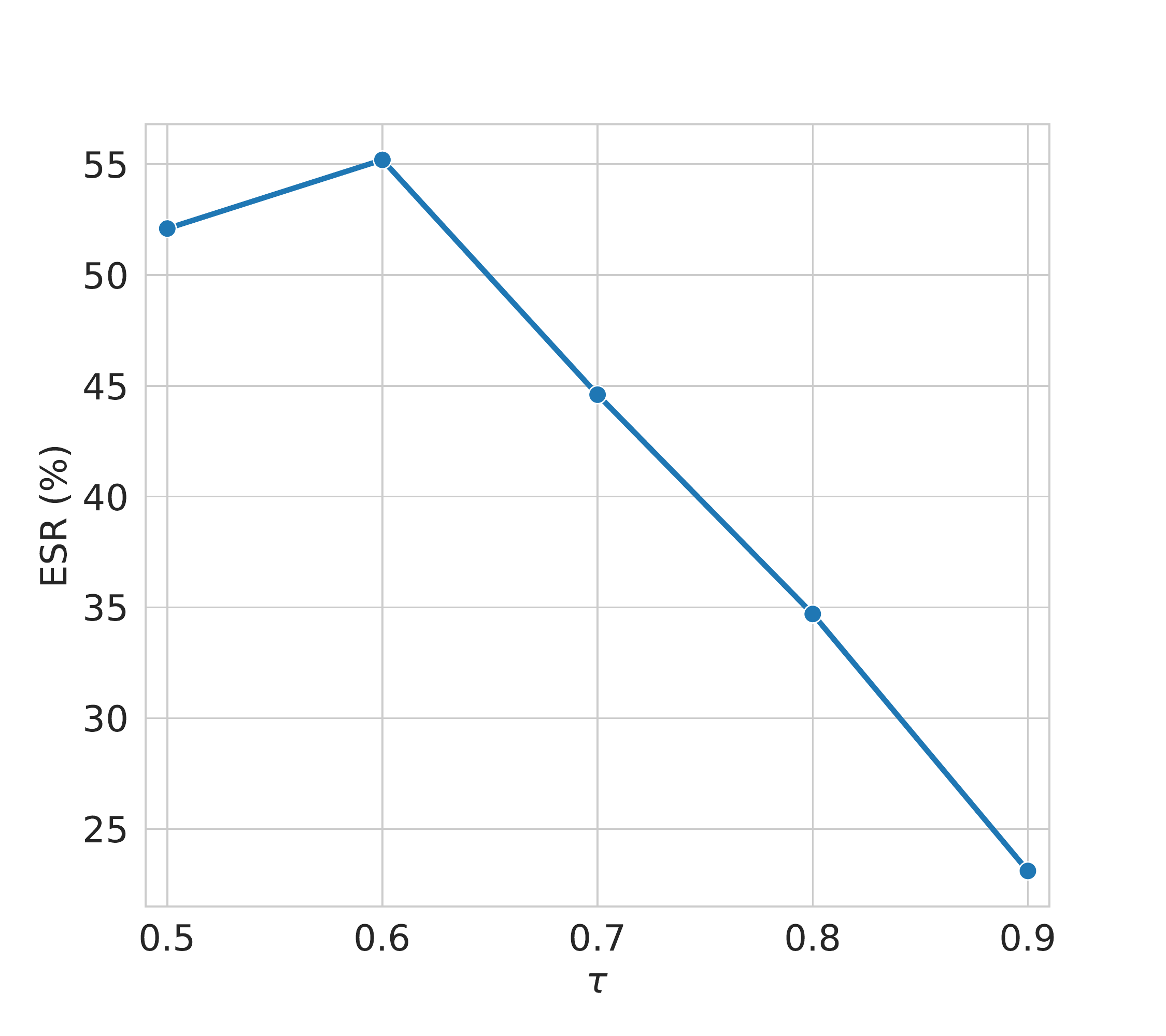}
		\caption{$\tau$}
		\label{fig-sensitivity-tau}
	\end{subfigure}
	\caption{Sensitivity on $T$, $\beta$, $\eta$ and $\tau$.}
	\label{fig-sensitivity}
\end{figure}

\textsf{FocusedCleaner}$_{CLD}$ has some vital hyperparameters which can influence its sanitation performance, i.e., temperature $T$ in the soft class probability, hyperparameter $\beta$ for the momentum trick, relative importance $\eta$ for the attribute smoother and the fraction $\tau$ for computing energy score's quantile. We present the sensitivity analysis on these four hyperparameters in Fig.~\ref{fig-sensitivity}. Fig.~\ref{fig-sensitivity-T} shows that crafting a suitable temperature for the GNN features will lead to better recovery performance, this is due to the increasing performance of the victim node detection module. Fig.~\ref{fig-sensitivity-beta} shows that choosing the appropriate combination of the quantile at $t$-th step and the threshold at ($t-1$)-th step will lead to a more suitable threshold for the unsupervised victim node detection. Fig.~\ref{fig-sensitivity-eta} demonstrates that introducing the feature smoothness penalty to the outer loss can also improve the sanitation performance of the \textsf{FocusedCleaner}. Fig.~\ref{fig-sensitivity-tau} verifies that setting $\tau=0.6$ is the best choice for the victim node detection. This phenomenon is rational since the too-narrow search space of the sanitizer will degenerate the sanitation performance. 

\subsection{Ablation Study}
The fact that \textsf{FocusedCleaner} outperforms GASOLINE-D demonstrates the importance of the victim node detection module as the ``focus" in sanitation. On the other hand, to show the importance of the bi-level structure learning module, we run the following experiments. Since the LinkPred-based detection module itself can detect adversarial links, we directly use it as a sanitizer without bi-level structure learning. We implement the \textsf{FocusedCleaner}$_{LP}$ and LinkPred on Cora dataset as an exemplar. Tab.~\ref{tab-ablation-bilevel} shows that \textsf{FocusedCleaner}$_{LP}$ outperforms LinkPred by a large margin. In combination, these results show that the structure learning and detection modules can enhance each other in the sanitation process. 
\begin{table}[h]
	\centering
	\caption{$\mathsf{ESR}$ (\%) for \textsf{FocusedCleaner}$_{LP}$ and LinkPred-based victim node detection.}
	\label{tab-ablation-bilevel}
	\resizebox{0.8\columnwidth}{!}{%
		\begin{tabular}{c|cc}
			\toprule[1.pt]
			Attacking power & \textsf{FocusedCleaner}$_{LP}$& LinkPred\\
			\hline
			$5\%$ &$29.2$&$12.5$\\
			$10\%$&$44.4$&$12.7$\\
			$15\%$&$49.0$&$13.5$\\
			$20\%$&$43.6$&$14.5$\\
			$25\%$&$37.8$&$13.7$\\
			\bottomrule[1.pt]
		\end{tabular}
	}
\end{table}

On the other hand, we also analyze the importance of the adaptive testing loss trick ($\lambda_{\mathcal{T}^{\prime}}\sum_{i\in\mathcal{T}^{\prime}}\sum_{c=1}^{C}\hat{\mathbf{y}}_{ic}\ln\mathbf{S}_{ic}$) for $\mathsf{FocusedCleaner}$. We set $\lambda_{\mathcal{V}}=1$ to represent the ignorance of the testing loss during sanitation, and the ablation study results are shown in Tab.~\ref{tab-ablation-CLD}. The results demonstrate that dynamically introducing the testing loss can better supervise the cleaner to assign a higher meta-gradient to the adversarial noise, thus leading to a better sanitation performance.
\begin{table}[h]
	\centering
	\caption{$\mathsf{ESR}$ (\%) for \textsf{FocusedCleaner}$_{CLD}$ with (w.) or without (w.o.) testing set $\mathcal{T}^{\prime}$ .}
	\label{tab-ablation-CLD}
	\resizebox{1.\columnwidth}{!}{%
		\begin{tabular}{l|rr}
			\toprule[1pt]
			Attacking power & \textsf{FocusedCleaner}$_{CLD}$-w.& \textsf{FocusedCleaner}$_{CLD}$-w.o.\\
			\hline
			$5\%$ &$36.0$&$33.7$\\
			$10\%$&$55.2$&$49.9$\\
			$15\%$&$52.3$&$48.0$\\
			$20\%$&$46.4$&$42.2$\\
			$25\%$&$39.7$&$37.6$\\
			\bottomrule[1pt]
		\end{tabular}
	}
\end{table}

Moreover, we also analyze the necessity of introducing ``focus" on the structure loss $\mathcal{L}_{S}$ by only considering the normal nodes in the validation and testing set. The results are shown in Tab.~\ref{tab-ablation-normal}. The experimental results demonstrate that introducing the ``focus" into the structure learning loss indeed boosts the sanitation performance of $\mathsf{FocusedCleaner}$.
\begin{table}[h]
	\centering
	\caption{$\mathsf{ESR}$ (\%) for \textsf{FocusedCleaner}$_{CLD}$ with (w.) or without (w.o.) normal set $\mathcal{N}$ as a ``focus" on the loss $\mathcal{L}_{S}$.}
	\label{tab-ablation-normal}
	\resizebox{1.\columnwidth}{!}{%
		\begin{tabular}{l|rr}
			\toprule[1pt]
			Attacking power & \textsf{FocusedCleaner}$_{CLD}$-w.& \textsf{FocusedCleaner}$_{CLD}$-w.o.\\
			\hline
			$5\%$ &$36.0$&$28.6$\\
			$10\%$&$55.2$&$46.9$\\
			$15\%$&$52.3$&$48.6$\\
			$20\%$&$46.4$&$42.0$\\
			$25\%$&$39.7$&$38.2$\\
			\bottomrule[1pt]
		\end{tabular}
	}
\end{table}

%\paragraph{Supplemental Experiment Results} 
%Additionally, we provide more comprehensive experiment results in the supplement, including sanitation results based on F1 score in Sec.~E, sensitivity analysis of hyperparameters in Sec.~F, ablation study of adaptive testing loss, bi-level structure learning module and normal node set as ``focus" for $\mathcal{L}_{S}$ in Sec.~G, and  ensembling two victim node detection methods in Sec.~H. 
%Combining FocusedCleaner with robust GNNs in Sec.~I.

\subsection{Sanitation for Robust Node Classification}
%\subsubsection{Results}
We proceed to evaluate \textsf{FocusedCleaner} as a preprocessing-based defense method and compare it to other defense approaches (both preprocessing-based and robust-model-based).
%we explore the robustness performance of \textsf{FocusedCleaner} with other recently state-of-the-art preprocessing-based methods and robust GNNs as an important application. The experiment results of combining \textsf{FocusedCleaner} with robust GNNs to defend against the poisoning attacks are shown in the appendix.
%To defend against structural poisoning attacks, there are two major types of defense approaches. The preprocessing-based approaches rely on sanitizing the poisoned graph first and feeding the sanitized graph into a vanilla GNN model. The robust-model-based approaches will design methods to train a robust model, which is then run over the poisoned graph. Below, we compare these two types of approaches.
%subsubsection{Preprocessing vs. Robust models}
%We investigate the two classes of approaches (pre-precessing based and robust model based)
%In this section, we start to analyze the robustness of different defense methods against two typical previous mentioned attacking methods. 
For preprocessing-based methods, we test GCN on the sanitized graph. For robust-model-based methods, we directly feed the poisoned graph to the robust models.
%\textit{NETTACK} \cite{nettack} to poison the graphs on the node classification task and use the preprocessing-based and robust-model-based methods separately as defense. 
Table~\ref{tab-defense-METTACK} and Table~\ref{tab-defense-MinMax}
summarize the node classification accuracies under different attack powers. Among all the methods, our proposed \textsf{FocusedCleaner}$_{CLD}$ and \textsf{FocusedCleaner}$_{LP}$ can achieve the best results in almost all cases, even outperforming robust-model-based methods. This shows that correctly eliminating the adversarial links might be a potentially better choice than mitigating the impairments of adversarial noises. 
%Moreover, within the type of preprocessing-based methods, \textsf{FocusedCleaner}$_{CLD}$ and \textsf{FocusedCleaner}$_{LP}$ perform better than state-of-the-art method GASOLINE. For example, on the Cora dataset with attacking powers  $20\%$ and $25\%$, \textsf{FocusedCleaner}$_{CLD}$ outperforms GASOLINE by $23.4\%$ and $28.2\%$ respectively. 
Besides, the consistency between the recovery ratio and the robustness performances shows that robustness is indeed a result of sanitation. 

We note that preprocessing-based and robust-model-based methods are two complementary defense approaches rather than conflicting. Indeed, we can feed the sanitized graph into the robust models, and we can observe a further performance boost. We show such results in the Sec.~I of the supplement. 

%In addition, we also test all the methods against another representative \textit{targeted} attack \textit{NETTACK} {\color{red} ref} (in comparison, \textit{METTACK} is a global untargeted attack). Even the sanitation methods are not designed for \textit{NETTACK}, we observe that \textbf{FocusedCleaner} outperforms others in most cases (in the supplement) in terms of defense performances, which demonstrates the potential of \textbf{FocusedCleaner} to generalize to other attacks. 

%For \textit{NETTACK}, our methods perform better than other methods in most of the cases, which shows that our framework can also effectively defend against targeted attacks. The reason of the performance of \textsf{FocusedCleaner} is suboptimal for Polblogs with a small perturbation rate is that the number of sanitation steps (sanitation ratio $\times$ poisoned graph links number) is far more than the true attack degree (target nodes number $\times$ Ptb/Node), thus resulting in the "over-learning" phenomenon to the cleaner. How to set the sanitation ratio for the preprocessing-based methods according to the attacking scenarios remains a problem to be solved in future work.

\subsection{\textsf{FocusedCleaner with Robust Models}}
The two types of defense approaches are not conflicting with each other but are rather complementary. That is, we can use preprocessing-based methods to sanitize the graph and then feed the graph into robust models. Specifically, we use MedianGNN and ElasticGNN as two representative robust models. Tab.~\ref{tab-defense-enhance-CLD} and Tab.~\ref{tab-defense-enhance-LP} show the node classification accuracies with/without the sanitation step. Indeed, sanitation will enhance the performance of robust models as sanitized graphs will have a higher level of homophily compared with poisoned graphs.

\begin{table}[h]
	\centering
	\caption{Mean accuracy (\%) for MedianGNN and ElasticGNN with (w.) or without (w.o.) sanitized graph by \textsf{FocusedCleaner}$_{CLD}$.}
	\label{tab-defense-enhance-CLD}
	\resizebox{1.\columnwidth}{!}{%
		\begin{tabular}{l|rrrrr}
			\toprule[1pt]
			Dataset& Attacking power & MedianGNN-w.& MedianGNN-w.o. & ElasticGNN-w. & ElasticGNN-w.o.\\
			\hline
			\multirow{5}*{Cora}&$5\%$ &$82.5$&$77.0$&$84.6$&$75.4$\\
			&$10\%$&$79.8$&$71.5$&$82.0$&$67.1$\\
			&$15\%$&$76.1$&$64.5$&$77.7$&$59.5$\\
			&$20\%$&$73.3$&$60.3$&$74.5$&$53.9$\\
			&$25\%$&$69.0$&$57.1$&$71.2$&$47.3$\\
			\hline
			\multirow{5}*{Citeseer}&$5\%$ &$74.6$&$64.0$&$75.3$&$65.1$\\
			&$10\%$&$72.3$&$58.6$&$72.5$&$58.1$\\
			&$15\%$&$66.8$&$55.1$&$67.0$&$52.4$\\
			&$20\%$&$63.1$&$50.7$&$62.1$&$47.6$\\
			&$25\%$&$59.8$&$47.2$&$58.3$&$42.3$\\
			\hline
			\multirow{5}*{Polblogs}&$5\%$ &$85.8$&$85.6$&$88.8$&$83.3$\\
			&$10\%$&$86.2$&$78.7$&$88.0$&$77.2$\\
			&$15\%$&$85.0$&$73.0$&$86.9$&$71.2$\\
			&$20\%$&$81.7$&$70.4$&$82.5$&$69.0$\\
			&$25\%$&$77.6$&$68.1$&$77.6$&$68.5$\\
			\bottomrule[1pt]                
		\end{tabular}
	}
\end{table}

\begin{table}[h]
	\centering
	\caption{Mean accuracy (\%) for MedianGNN and ElasticGNN with (w.) or without (w.o.) sanitized graph by \textsf{FocusedCleaner}$_{LP}$.}
	\label{tab-defense-enhance-LP}
	\resizebox{1.\columnwidth}{!}{%
		\begin{tabular}{l|rrrrr}
			\toprule[1pt]
			Dataset& Ptb rate (\%) & MedianGNN-w.& MedianGNN-w.o. & ElasticGNN-w. & ElasticGNN-w.o.\\
			\hline
			\multirow{5}*{Cora}&$5\%$ &$80.2$&$77.0$&$82.0$&$75.4$\\
			&$10\%$&$78.9$&$71.5$&$80.7$&$67.1$\\
			&$15\%$&$76.6$&$64.5$&$78.2$&$59.5$\\
			&$20\%$&$71.7$&$60.5$&$73.1$&$53.9$\\
			&$25\%$&$68.4$&$57.1$&$68.8$&$47.3$\\
			\hline
			\multirow{5}*{Citeseer}&$5\%$ &$73.4$&$64.0$&$73.6$&$65.1$\\
			&$10\%$&$71.1$&$58.6$&$70.8$&$58.1$\\
			&$15\%$&$67.2$&$55.1$&$66.0$&$52.4$\\
			&$20\%$&$63.7$&$50.7$&$63.5$&$47.6$\\
			&$25\%$&$61.2$&$47.2$&$61.1$&$42.3$\\
			\hline
			\multirow{5}*{Polblogs}&$5\%$ &$86.7$&$85.6$&$86.3$&$83.3$\\
			&$10\%$&$84.9$&$78.7$&$85.3$&$77.2$\\
			&$15\%$&$81.7$&$73.0$&$81.3$&$71.2$\\
			&$20\%$&$77.5$&$70.4$&$78.0$&$69.0$\\
			&$25\%$&$74.7$&$68.1$&$76.3$&$68.5$\\
			\bottomrule[1pt]             
		\end{tabular}
	}
\end{table}
\section{Conclusion}
In this paper, we target on the graph sanitation problem which aims at pruning the maliciously adversarial links from the poisoned graph. In particular, we propose a new framework--\textsf{FocusedCleaner} to joint learn a bi-level structure learning module with a crafted unsupervised victim node detection module. Specially, the victim node detection module provide the ``focus" to help the structure learning module precisely identify and prune the candidate adversarial links. To validate the sanitation performance, we design a new metric--$\mathsf{ESR}$ based on the Jaccard index to scientifically measure the sanitation quality. The experimental results demonstrate that our sanitizers achieve comparable performances both on graph sanitation and graph defense task. 

%In this paper, we rethink the conventional graph sanitation problem, which prompts us to sanitize the adversarial noises from the Web environment rather than ``neutralize” the contamination provided by malicious users. From this perspective, we propose a joint learning framework to supervise the bi-level structural learning module with the crafted unsupervised victim node detections' prediction to precisely identify and sanitize the candidate adversarial noises of the Web platform. {\color{red} Since some metrics in the victim node detection are time-consuming, we reformulate and vectorize them in an efficient version and feed them into the deep Gaussian mixture model with an adaptive thresholding strategy for more precise anomaly detection performance -- these things are too minor compared to others}. On the other hand, we also explore the GNN-based link prediction model as the victim node detector and verify its effectiveness. At last, we design the recovery ratio $\mathcal{R}$ and explore the sanitation accuracy and robustness performance of the state-of-the-art preprocessing-based methods and robust GNNs and prove the strength of \textsf{FocusedCleaner}.

\bibliographystyle{ieeetr}
\bibliography{citation}

\end{document}